\documentclass[journal]{IEEEtran}
\usepackage{amssymb}
\usepackage{amsmath}
\usepackage{epsfig}
\usepackage{pmat}
\usepackage[noend]{algpseudocode}
\usepackage{algorithmicx,algorithm}
\usepackage{multirow}
\usepackage{diagbox}[2011/11/22]
\usepackage{booktabs}
\usepackage{color}
\usepackage{threeparttable}

\ifCLASSINFOpdf
\else
\fi

\begin{document}
%
\title{Polarimetric Convolutional Network for PolSAR Image Classification}
%
%

\author{Xu~Liu,~\IEEEmembership{Student~Member,~IEEE,}
Licheng~Jiao,~\IEEEmembership{Fellow,~IEEE,}
Xu~Tang,~\IEEEmembership{Member,~IEEE,}
Qigong~Sun,~\IEEEmembership{Student~Member,~IEEE,}
Dan~Zhang
\thanks{This work was supported in part by the Major Research Plan of the National Natural Science Foundation of China (No. 91438201 and No. 91438103),
the National Natural Science Foundation of China (No. 61801351)
the National Science Basic Research Plan in Shaanxi Province of China (No.2018JQ6018),
the Fund for Foreign Scholars in University Research and Teaching Programs (the 111 Project) (No. B07048), the Fundamental Research Funds for the Central Universities ( No. XJS17108) and the China Postdoctoral Fund (No. 2017M613081).}

\thanks{The authors are with  the Key Laboratory of Intelligent Perception and Image Understanding of the Ministry of Education of China, International Research Center of Intelligent Perception and Computation,
School of Artificial Intelligence, Xidian University, Xi'an, Shaanxi Province 710071, China (e-mail: xuliu361@163.com; lchjiao@mail.xidian.edu.cn). }}

\maketitle

\begin{abstract}
The approaches for analyzing the polarimetric scattering matrix of polarimetric synthetic aperture radar (PolSAR) data have always been the focus of PolSAR image classification.
Generally, the polarization coherent matrix and the covariance matrix obtained by the polarimetric scattering matrix are used as the main research object to extract features.
In this paper, we focus on the original polarimetric scattering matrix and propose a polarimetric scattering coding way to deal with polarimetric scattering matrix and obtain a close complete feature. This encoding mode can also maintain polarimetric information of scattering matrix completely.
At the same time, in view of this encoding way, we design a corresponding classification algorithm based on convolution network to combine this feature. Based on polarimetric scattering coding and convolution neural network, the polarimetric convolutional network is proposed to classify PolSAR images by making full use of polarimetric information.
We perform the experiments on the PolSAR images acquired by AIRSAR and RADARSAT-2 to verify the proposed method. The experimental results demonstrate that the proposed method get better results and has huge potential for PolSAR data classification. Source code for polarimetric scattering coding is available at \emph{https://github.com/liuxuvip/Polarimetric-Scattering-Coding}.
\end{abstract}

\IEEEpeerreviewmaketitle

\section{Introduction}
%
%
%
%
\label{Intr}
Polarimetric synthetic aperture radar (PolSAR) images collected with airborne and satellite sensors are a wealthy source of information concerning the Earth's surface, and have been widely used in urban planning, agriculture assessment and environment monitoring \cite{wang2017comparison, voormansik2016observations}. These applications require the fully understanding and interpretation of PolSAR images.

Hence, PolSAR image interpretation is of much significance in theory and application. Land use classification of PolSAR images are an important and indispensable research topic since these images contain rich character of the target (e.g., scattering properties, geometric shapes, and the direction of arrival). The land use classification is arranging the pixels to the different categories according to the certain rule. The common objects within the PolSAR images include land, buildings, water, sand, urban areas, vegetation, road, bridge and so on \cite{liuf2016hierarchical}.
In order to distinguish them, the features of the pixels should be fully extracted and mined.
With the development of the PolSAR image classification, many feature extraction algorithms  based on physical scattering mechanisms have been introduced.
The feature extraction techniques based on polarimetric characteristics can be divided into two kinds: coherent target decomposition and incoherent target decomposition. The former acts on the scattering matrix to characterize completely polarized scattered waves, which contains the fully polarimetric information. The latter acts only on the mueller matrix, covariance matrix, or coherency matrix in order to characterize partially polarized waves \cite{dickinson2013classification}.

The coherent target decomposition algorithms include the Pauli decomposition, the sphere-diplane-helix (SDH) decomposition \cite{Krogager1990New}, the symmetric scattering characterization method (SSCM) \cite{Touzi2002Characterization}, Cameron decomposition \cite{Cameron1990Feature}, Yamaguchi Four-component scattering power decomposition\cite{Yamaguchi2011Four}, General polarimetric model-based decomposition \cite{Chen2013General, Chen2013Adaptive}, and some advances \cite{Chen2018Advanced, Mahdianpari2018Fisher}.
The incoherent target decomposition algorithms include Huynen decomposition \cite{Huynen1978}, Freeman-Durden decomposition \cite{Freeman1993Three}, Yamguchi four-component decomposition \cite{Yamaguchi2005Four}, Cloude-Pottier decomposition \cite{Cloude1996A}, \cite{Cloude1997An}, and a
number of approaches have been reported\cite{Lee2014Generalized, Besic2014Polarimetric, Aghababaee2016Incoherent}.
In addition to feature based on the polarization mechanism \cite{Chen2014Modeling, Chen2014Uniform, Xu2017Polarimetric, Tao2017PolSAR},
there are some traditional features of natural images, which have been utilized to analyze PolSAR image, such as color features \cite{Uhlmann2014Integrating}, texture features \cite{Grandi2007Target}, spatial relations \cite{Ma2014Polarimetric}, etc.
Based on the above basic features, some multiple features of PolSAR data have been constructed to improve the classification performance \cite{Zou2010Polarimetric, Uhlmann2014Integrating, ren2017unsupervised}.

For classification tasks, besides the feature extraction, classifier design is also a key point. According to the degree of data mark, the classification methods can be broadly divided into three groups, including unsupervised classification (without any labeled training data), semi-supervised classification (SSC) (with a small amount of labeled data and a large amount of unlabeled data) and supervised classification (with completely labeled training data) \cite{liu2016semisupervised}.

The unsupervised classification approaches design a function to describe hidden structure from unlabeled data.
The traditional methods always make a decision rule to cluster PolSAR data into different groups, and the number of groups is also a hyper-parameter. There are a lot of unsupervised classification methods for PolSAR data, such as H/$\alpha$ complex Wishart classifier \cite{Lee2002Unsupervised}, polarimetric scattering characteristics preserved method \cite{Lee2004Unsupervised}, Fuzzy k-means cluster classifier \cite{L1996Fuzzy, Kersten2005Unsupervised}, the classification based on deep  learning \cite{Liu2016POL,Jiao2016Wishart,LCJDeep}, etc.
The SSC is a class of supervised learning tasks and techniques that also make use of a small amount of labeled data and a large amount of unlabeled data for training. Compared with the unsupervised method, SSC can improve classification performance so long as the target is to make full use of a smaller number of labeled samples \cite{Liu2016Large}.
There are many semi-supervised classification for PolSAR data, such as the classification based on hypergraph learning \cite{Wei2014PolSAR},
 the method based on parallel auction graph \cite{Liu2016Fast}, spatial-anchor graph \cite{Liu2016Large}, etc.
Unlike unsupervised approaches and semi-supervised approaches, the supervised classifications use enough labeled samples to train the classifiers which can be applied to determine the class of other samples. Lots of methods have been introduced, including maximum likelihood \cite{Harant2010Fisher},
support vector machines \cite{Fukuda2001Polarimetric, Li2008Object, Aghababaee2013Contextual}, sparse representation \cite{Zhang2015Fully}, deep learning \cite{Li2009Improving, Zhou2016Polarimetric, Liu2017Polarimetric}.

Recently, deep learning has attracted considerable attention in the computer vision community \cite{Krizhevsky2012ImageNet,Bengio2009Learning,Yang2015Learning,Zhang2013Tensor},
as it provides an efficient way to learn image features and to represent certain function classes far more efficiently than shallow ones \cite{Chen2014Deep}, \cite{Han2015Object}, \cite{TDC7529190}.
Deep learning has also been introduced into the geoscience and remote sensing (RS) community \cite{Liu2016POL}, \cite{gong2016change}, \cite{Zhao2016Spectral}, \cite{Petersson2017Hyperspectral}, \cite{Yu2017Convolutional, XuTwo, XL8048556}, \cite{lu2017remote,lu2015semi}, \cite{yang2018deep}. 
Especially in the direction of PolSAR image classification,
in \cite{Jiao2016Wishart}, a specific deep model for polarimetric synthetic aperture radar (POLSAR) image classification is proposed, which is named as Wishart deep stacking network (W-DSN). A fast implementation of Wishart distance is achieved by a special linear transformation, which speeds up the classification of POLSAR image.
In \cite{Liu2016POL}, a new type of restricted boltzmann machine (RBM) is specially defined, which we name the Wishart-Bernoulli RBM (WBRBM), and is used to form a deep network named as Wishart Deep Belief Networks (W-DBN).
In \cite{xie2017polsar}, a new type of autoencoder (AE) and convolutional autoencoder (CAE) is specially defined, which we name them Wishart-AE (WAE) and Wishart-CAE (WCAE).
In \cite{zhang2017complex}, a complex-valued CNN (CV-CNN) specifically for synthetic aperture radar (SAR) image interpretation. It utilizes both amplitude and phase information of complex SAR imagery.
In \cite{Chen2018PolSAR}, Si-Wei Chen, Xue-Song Wang and Motoyuki Sato proposed a polarimetric-feature-driven deep convolutional neural network (PFDCN) for PolSAR image classification. The core idea of which is to incorporate expert knowledge of target scattering mechanism interpretation and polarimetric feature mining to assist deep CNN classifier training and improve the final classification performance.

What is more, fullly convolutional network (FCN) is successfully used for natural image semantic segmentation  \cite{Badrinarayanan2017SegNet, Audebert2016Semantic,shelhamer2017fully, siam2018rtseg, tsai2018learning, liang2018dynamic, chen2018deeplab} and remote sense image classification based on one by one pixel \cite{isikdogan2017surface, cheng2017automatic, jiao2017deep, volpi2017dense}.
In \cite{isikdogan2017surface}, a fully convolutional neural network is trained to segment water on Landsat imagery. In \cite{cheng2017automatic}, a novel deep model, i.e.,a cascaded end-to-end convolutional neural network (CasNet), was proposed to simultaneously cope with the road detection and centerline extraction tasks. Specifically, CasNet consists of two networks.
One aims at the road detection task. The other is cascaded to the former one, making full use of the feature maps produced formerly, to obtain the good centerline extraction. In \cite{jiao2017deep}, a novel hyperspectral image classification (HSIC) framework, named deep multiscale spatial-spectral feature extraction algorithm, was proposed based on fully convolutional neural network. In \cite{volpi2017dense}, the authors presented a CNN-based system relying on a downsample-then-upsample architecture. Specifically, it first learns a rough spatial map of high-level representations by means of convolutions and then learns to upsample them back to the original resolution by deconvolution. By doing so, the CNN learns to densely label every pixel at the original resolution of the image.
%

Inspired by the previous research works, a supervised PolSAR image classification method based on polarimetric scattering coding and convolution network is proposed in this paper.
Our goal is to solve the problems of PolSAR data coding, feature extraction, and land cover classification.
Our work can be summarized into three main parts as follows.
\begin{itemize}
  \item Firstly, a new encoding mode of polarimetric scattering matrix is proposed, which is called polarimetric scattering coding. It not only completely preserves the polarization information of data, but also facilitates to extract high-level features by deep learning, especially convolutional networks.
  \item Secondly, a novel PolSAR image classification algorithm based on polarimetric scattering coding and convolutional network is proposed, which is called polarimetric convolutional network and also an end-to-end learning framework.
  \item Thirdly, feature aggregation is designed to fuse the two kinds of feature and mine more advanced features.
\end{itemize}

The paper is organized as follows. In Section \uppercase\expandafter{\romannumeral2},
the representation of PolSAR images is described.
 In Section \uppercase\expandafter{\romannumeral3}, the proposed method named polarimetric convolutional network is given.
 The experimental setting is presented in Section \uppercase\expandafter{\romannumeral4}. The results are presented in Section \uppercase\expandafter{\romannumeral5}, followed by the conclusions and discussions in Section \uppercase\expandafter{\romannumeral6}.

\section{Proposed method}
\label{Prmd}
In the PolSAR image classification task, the land use classes are determined by different analysis include polarization of the target responses, scattering heterogeneity determination and determination of the polarization state for target discrimination, which need to be decided by different features. It is hard for researchers to consider all kinds of features.
PolSAR data is a two-dimensional complex matrix. The traditional feature extraction method represents PolSAR data into a one-dimensional vector, which destroys data space structures.
In order to solve this problem, the intuitive way is to express the original data directly. In this paper, the polarimetric scattering coding is proposed to express the original data directly, which can maintain structure information completely. Next, the polarimetric scattering coding matrix obtained by the encoding is fed into a classifier based on fully convolutional network.
The following of this section consists of three parts. First, representation of PolSAR images is given. Second, the polarimetric scattering coding for complex scattering matrix $S$ is explained. Third, the proposed method called polarimetric convolutional network is presented.
\subsection{Representation of PolSAR images}
The fully PolSAR measures the amplitudes and phases of backscattering signals in four combinations: 1) HH; 2) HV; 3) VH; and 4) VV. Where H means horizontal mode, V means vertical mode. These signals form a $2 \times 2$ complex scattering matrix $S$ to represent the information for one pixel, which relates the incident and the scattered electric fields.
Scattering matrix $S$ can be expressed as

\begin{equation}
\label{eqs1}
S=
\begin{bmatrix}
S_{HH} & S_{HV} \\
S_{VH} & S_{VV}
\end{bmatrix}
=
\begin{bmatrix}
\left|S_{HH}\right|{e}^{i\times\phi_{HH}} & \left|S_{HV}\right|{e}^{i\times\phi_{ HV}}  \\
\left|S_{VH}\right|{e}^{i\times\phi_{VH}}  & \left|S_{VV}\right|{e}^{i\times\phi_{VV}}
\end{bmatrix}
\end{equation}

where $S_{HH},S_{HV},S_{VH}$ and $S_{VV}$ are the complex scattering coefficients, $S_{HV}$ is the scattering coefficient of the horizontal(H)
transmitting and vertical(V) receiving polarization. Other elements have similar definitions.
$\left|S_{HH}\right|$, $\left|S_{HV}\right|$, $\left|S_{VH}\right|$, $\left|S_{VV}\right|$ denote the amplitudes of the measured complex scattering coefficients,
$\phi_{HH}$, $\phi_{HV}$, $\phi_{VH}$ and $\phi_{VV}$ are the value of phases. $i$ is the complex unity.

The characteristics of the target can be specified by vectorizing the scattering matrix. Based on two important basis sets, lexicographic basis and Pauli spin matrix set, in the case of monostatic backscattering with reciprocal medium, the lexicographic scattering vector $\vec{k}_{L}$ and Pauli scattering vector $\vec{k}_{p}$ are defined  as
\begin{align}
\vec{k}_{L} &= [S_{HH}, \sqrt{2}S_{HV}, S_{VV}]^{T}\\
\vec{k}_{p} &= 1/\sqrt{2}[S_{HH}+S_{VV}, S_{HH}-S_{VV},2S_{HV}]^{T}
\end{align}
where superscript $T$ denotes the transpose of vector.

The scattering characteristics of a complex target are determined by different independent subscatterers and their interaction, The scattering characteristics described by a statistic method due to the randomness and depolarization. Moreover, the inherent speckle in the SAR data reduced by spatial averaging at the expense of lossing spatial resolution. Therefore, for the complex target, the scattering characteristics should be described by statistic coherence matrix or covariance matrix. Covariance and coherence matrices can be generated from the outer product of $\vec{k}_{L}$ and $\vec{k}_{p}$ respectively, with its conjugate transpose
\begin{align}
\textbf{C}&=<\vec{k}_{L}\vec{k}_{L}^{\textbf{H}}>\\
\textbf{T}&=<\vec{k}_{P}\vec{k}_{P}^{\textbf{H}}>
\end{align}
where $<\cdot >$ denotes the average value in the data processing stage, and the superscript $\textbf{H}$ stands for the complex conjugate and transpose of vector and matrix.

The covariance matrix $\textbf{C}$ has been proved to follow a complex
Wishart distribution \cite{Lee2009Polarimetric}. Moreover, the coherence matrix $\textbf{T}$ is used to express PolSAR data, which has a linear relation with covariance matrix $\textbf{C}$.
The PolSAR features always extracted indirectly from the PolSAR data, such as color features,
texture features, and the decomposition features. The color and texture features are extracted from the pseudo-color image which is comprised of the decomposition components. The spatial relation of the pixels could be obtained from such a PolSAR pseudo-color image. The decomposition features are made up of matrix $\textbf{C}$ or $\textbf{T}$ by polarimetric target decompositions, e.g., Pauli decomposition, Cloude decomposition, Freeman-Durden decomposition, etc. A number of works have included the computations of these features as shown in \cite{De2007Target, Uhlmann2014Integrating}.

\subsection{Polarimetric scattering coding}
\label{ssc}
Polarimetric data returned by polarimetric synthetic aperture radar is stored in the polarimetric scattering matrix.
Polarimetric scattering matrix is used to show polarimetric information, in which the element is complex value.
Inspired by the one-hot coding \cite{harris2010digital} and hash coding \cite{leskovec2014mining}, we learned from the idea of position encoding and mapping relationship, proposed the polarimetric scattering coding for complex matrix encoding.
We assume that $z=(x+yi)$
is a complex value,
$x$ and $y$ are the real and imaginary parts of $z$ respectively. Considering the sign of $x$ and $y$, there are four possible for $z$, we give a complete encoding as follows, which is called sparse scattering coding $\varphi$ by us.

\begin{equation}
\label{Newe}
\varphi \left ( x+yi \right ) = \left\{
              \begin{array}{rrr}
                    \begin{bmatrix}
                    x & 0 \\
                    0 & \left | y \right |
                    \end{bmatrix},
                    if \; x\ge0 \; and \; y\le0\\
\specialrule{0em}{0.5ex}{0.5ex}
                    \begin{bmatrix}
                    x & y \\
                    0 & 0
                    \end{bmatrix},
                    if \; x\ge0 \; and \; y>0\\
\specialrule{0em}{0.5ex}{0.5ex}
                    \begin{bmatrix}
                    0 & y \\
                    \left | x \right | & 0
                    \end{bmatrix},
                    if \; x<0 \; and \; y>0\\
\specialrule{0em}{0.5ex}{0.5ex}
                    \begin{bmatrix}
                    0 & 0 \\
                    \left | x \right | & \left | y \right |
                    \end{bmatrix},
                    if \; x<0 \; and \; y\le0\\
              \end{array}
 \right.
\end{equation}

\begin{figure}[htb]
  \centering
\begin{minipage}[b]{0.48\linewidth}
  \centering
\centerline{\hspace{0.7cm}\epsfig{figure= 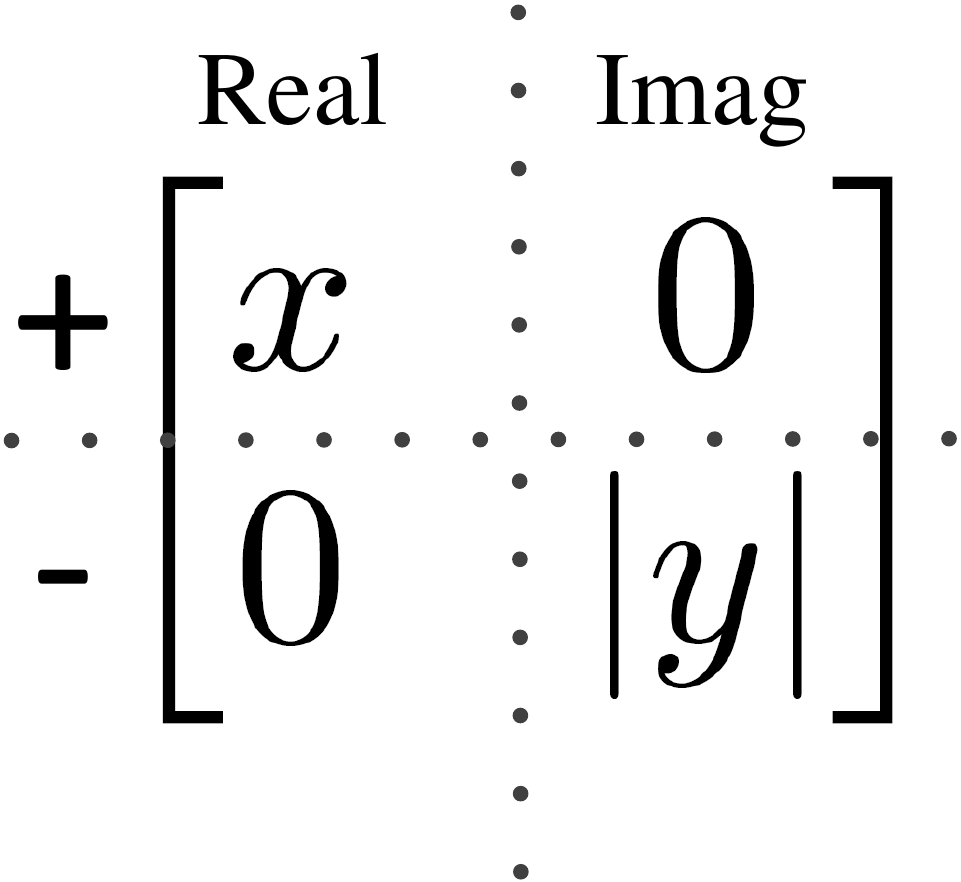,width=3.0cm}}
  \vspace{0.25cm}
\end{minipage}
\caption{The polarimetric scattering coding graphic of Eq. (\ref{Newe}). When $x>0, y<0$.}
\label{s1}
\end{figure}
Fig. \ref{s1} and Eq. (\ref{Newe}) show the details of polarimetric scattering coding,
the first column represents the real element,
the second column represents the imaginary part of the element,
the first line represents the positive element,
and negative elements are expressed in the second row. The $\left | \cdot  \right |$ is the absolute value operation.
An example in Eq. (\ref{Newe}) is given as follows

\begin{equation}
\varphi \left ( x+yi \right ) =
\begin{bmatrix}
x & 0 \\
0 & \left | y \right |
\end{bmatrix},
if \; x\ge0 \; and \; y<0\\
\end{equation}

$\varphi$ represents the function of polarimetric scattering coding, when $x>0$, $y<0$. From Eq. (\ref{eqs1}), scattering matrix $S$ can been written as
\begin{equation}
S=
\begin{bmatrix}
S_{HH} & S_{HV} \\
S_{VH} & S_{VV}
\end{bmatrix}
\end{equation}

\begin{figure}[htb]
  \centering
\begin{minipage}[b]{0.1\linewidth}
  \centering
\centerline{\hspace{0.1cm}\epsfig{figure= 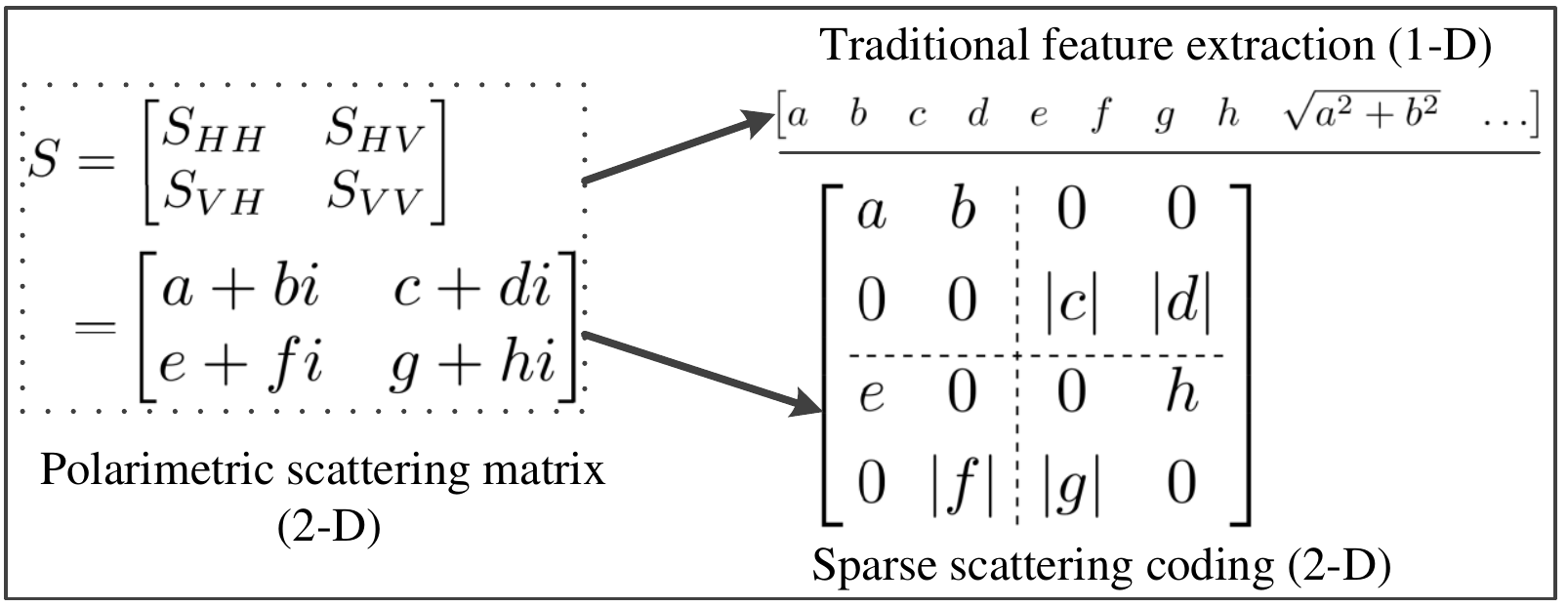,width=8.9cm}}
  \vspace{0.25cm}
\end{minipage}
\caption{The difference of traditional feature extraction and polarimetric scattering coding for one pixel in the PolSAR image.}
\label{explct}
\end{figure}

Because $S$ is a complex matrix, we can write its elements as follows
\begin{align}
S_{HH} &= a+bi \\
S_{HV} &= c+di \\
S_{VH} &= e+fi \\
S_{VV} &= g+hi
\end{align}
 In order to facilitate the understanding and explanation, we give a general assumption, where $a, b, e, h > 0$, $c, d, f, g < 0$. This assumption can take into account the characteristics of the PolSAR data. For instance, some PolSAR data format is int16 (-32,768 to +32,767). Polarimetric scattering coding of the scattering matrix $S$ can be given, which is called polarimetric scattering coding matrix.
\begin{align}
\nonumber \varphi \left (S \right ) &= \varphi \Bigg(\begin{bmatrix}
S_{HH} & S_{HV} \\
S_{VH} & S_{VV}
\end{bmatrix}\Bigg) \\
\nonumber &= \varphi \Bigg(
\begin{bmatrix}
a+bi & c+di \\
e+fi & g+hi
\end{bmatrix}\Bigg) \\
&= \begin{pmat}[{.|}]
a & b                  & 0                  & 0                 \cr
0 & 0                  & \left | c \right | & \left | d \right | \cr \-
e & 0                  & 0                  & h \cr
0 & \left | f \right | & \left | g \right | & 0 \cr
  \end{pmat}
\end{align}
Based on this new coding way, we can get the polarimetric scattering coding matrix, which is a 2-D sparse matrix. We also avoid transforming a complex matrix into a 1-D vector, as is shown in Fig \ref{explct}.
This will bring great convenience to the processing of PolSAR data.
The proposed polarimetric coding could show more information and is easy to generate and restore polarimetric covariance matrix.
Next to it, we proposed a new corresponding classification method named polarimetric convolutional network.

\begin{figure*}[htb]
  \centering
\begin{minipage}[b]{0.48\linewidth}
  \centering
\centerline{\hspace{0.7cm}\epsfig{figure= 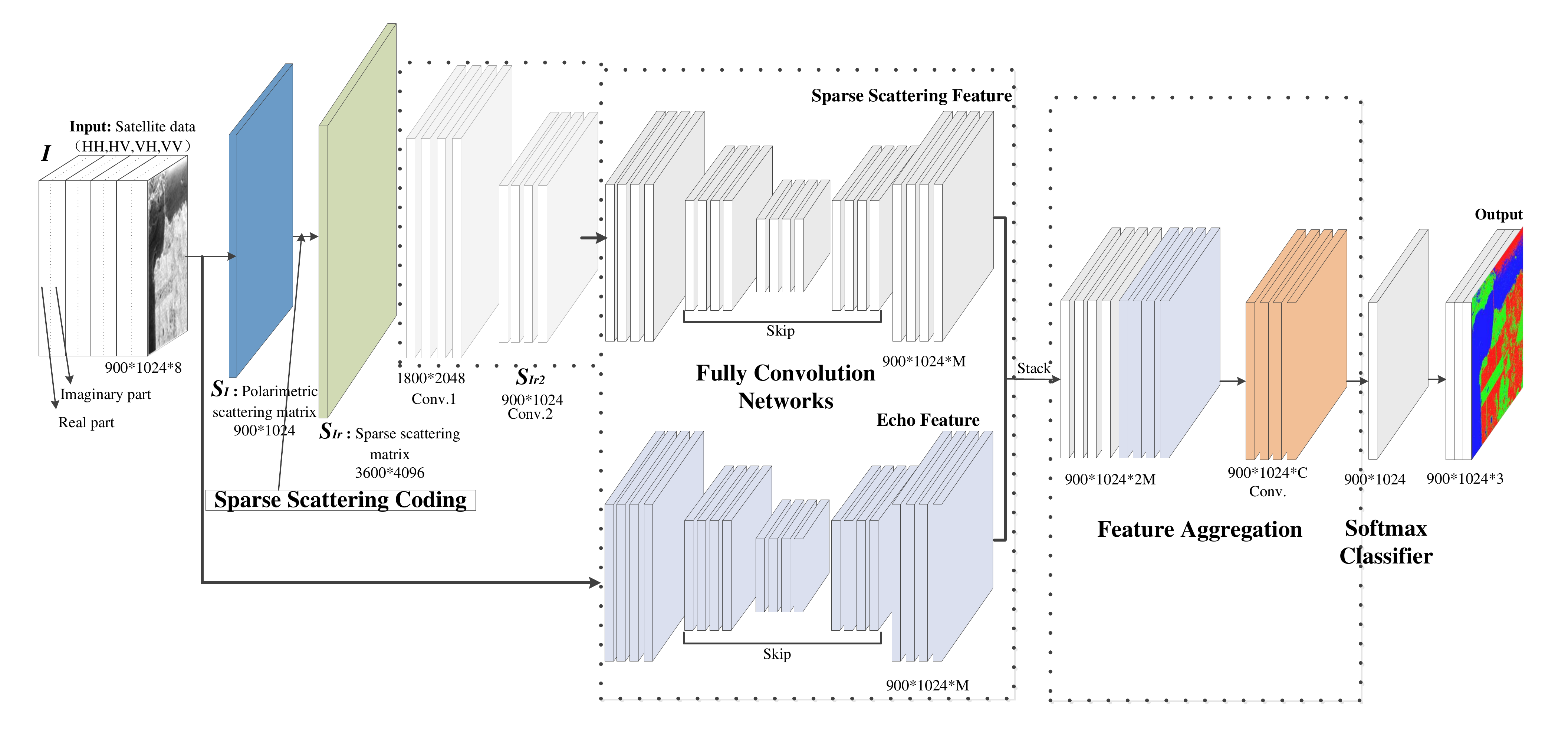,width=14.0cm}}
  \vspace{0.25cm}
\end{minipage}
\caption{The flowchart of polarimetric convolutional network. There are 8 channels in the original input data $I$, which contains four modes, and each has real and imaginary parts. For a pixel, eight channel elements can be represented by a-b in Formula. \ref{eqs1}, and the input data $I$ is encoded into a complex value matrix $S_{I}$, called polarimetric scattering matrix. By polarimetric scattering coding, polarimetric scattering coding matrix $S_{Ir}$ can been obtained. The feature map $S_{Ir2}$ is generated from the two layer convolutional networks. $S_{I}$ and $S_{Ir2}$
 are fed into two fully convolutional networks to get sparse scattering feature and echo feature, respectively.
 Finally, two kinds of feature from the two networks are aggregated to get the classification results.}
\label{flowchart}
\end{figure*}

\begin{algorithm}[htb]
\caption{Polarimetric scattering coding}
\hspace*{0.02in} {\bf Input:}
The raw polarimetric data\\
\hspace*{0.02in} {\bf Output:}
Polarimetric scattering coding matrix;
\begin{algorithmic}[1]
\State Calculate polarimetric data and get scattering matrix $S$
\State Polarimetric scattering coding $S$ and get polarimetric scattering coding matrix $\varphi(S)$
\State \Return polarimetric scattering coding matrix
\end{algorithmic}
\end{algorithm}

\subsection{Polarimetric convolutional network}
\label{PCN}
Recently, convolutional network performs well in the task of image semantic segmentation, especially fully convolutional network (FCN).
Actually, compared to other models, FCN can not only predict class attributes,
but also infer the spatial location and contour information,
which are helpful to the pixel level classification task.
Differences with conventional convolutional network, FCN replaces the fully connected layers with convolutional layers to output spatial maps instead of classification scores, those maps are upsampled using deconvolution to obtain dense per-pixel labeled outputs.
The components of FCN mainly includes convolutional layer (encode) and deconvolutional layer (decode).
In the forward propagation stage, the three-dimensional input data was
down-sampled in the low convolutional layers with relatively convolutional filters. The intermediate feature maps were up-sampled in the high deconvolutional layers with corresponding filters. At last, a softmax classifier followed to predict pixel-wise labels for an output map which has the same size as the input image. The output of the softmax classifier is a $C$ channels map of probabilities where $C$ is the number of classes.
In the back propagation learning stage, stochastic gradient descent was used to calculate parameters according to the difference between predictions and ground truth maps.

\begin{algorithm}[t]
\caption{Polarimetric convolutional network}
\hspace*{0.02in} {\bf Input:}
The raw polarimetric data $I$\\
\hspace*{0.02in} {\bf Output:}
Classification map;
\begin{algorithmic}[1]
\State Calculate $I$ and get $S_{I}$
\State Polarimetric scattering coding $S_{I}$ and get $S_{Ir}$
\State Input $S_{Ir}$ to a two layer convnet and get $S_{Ir2}$
\State Input $I$ and $S_{Ir2}$ to two different FCNs individually, and get two feature maps.
\State Stack and aggregate the two feature maps.
\State Classify the aggregated feature by softmax classifier
\State Update 3-6 by stochastic gradient descent
\State \Return classification map
\end{algorithmic}
\end{algorithm}

In this paper, a polarimetric convolutional network is proposed for polarimetric SAR data classification, as is shown in Fig. \ref{flowchart}.
First, the raw polarimetric data $I$ can be equivalently converted into polarimetric scattering matrix $S_{I}$, the size of $I$ is $s1 \times s2 \times 8$, the size of $S_{I}$ is $s1 \times s2$, which is a complex value matrix.
Second, polarimetric scattering matrix can be transformed from complex value matrix to real value matrix  $S_{Ir}=\varphi(S_{I})$ by polarimetric scattering coding, the size of $S_{Ir}$ is $4s1\times 4s2$.
Third, the spare scattering matrix $S_{Ir}$ is fed into a two layers convolutional networks and get feature maps $S_{Ir2}$ with a size of $s1 \times s2$.
Fourth, the feature maps generated from the two-layer convolutional networks can be fed into a fully convolutional network whose input size is same as the output size.
Fifth, another pathway is also a fully convolutional network, the input data is the the raw polarimetric data $I$.
Sixth, the part of feature aggregation is designed to fuse the two kinds of the feature. The last layer feature maps of the two pathway are stacked together as new feature maps, the size of the new feature maps is $s1 \times s2 \times (2\times M)$, $M$ is the number of feature maps in each pathway.
At last, in order to use softmax to classify data into $C$ classes, we add a convolutional layer to reduce the dimension of the new feature maps from $(2\times M)$ to $C$.

Above all, The whole network is an end-to-end network. For the entire network architecture, the loss function is a sum over the spatial dimensions of the final layer. We can use stochastic gradient descent to optimize it.

In the proposed method, the model mainly includes polarimetric scattering coding layer, convolutional layer, and deconvolutional layer. The details of polarimetric scattering coding layer are in Section. \ref{ssc}.
The convolutional layer can be defined as below.
\begin{align}
O_{ij} = F\left [ \sum^{k}_{m_{i},m_{j}=0} G(X_{s_{i}+m_{i},s_{j}+m_{j}})\right ]
\end{align}
where $O_{ij}$ is the output map of the convolutional layer in $i$ row and $j$ column, $F$ denotes the normalized function, rectified linear unit (ReLU) is a good choice and used in this article. $G$ is the convolutional function, $k$ is the size of the kernel, $X$ denotes the input pixels from the form layer, $s$ denotes the sampling stride.

The deconvolution layer is composed of upsampling and convolutional layers, upsampling layer corresponds to a max-pooling one in the downsampling stage. Those layers upsample feature maps using the max-pooling indices from their corresponding feature maps. The upsampled maps are then convolved with a set of trainable convolutional kernels to produce dense feature maps.

At last, we compute the energy function by a pixel-wise softmax over the final feature maps and the cross entropy loss function. The softmax is defined as

\begin{align}
p_{c}(x) = \frac{\exp(a_{c}(x))}{\sum ^{C}_{c'=1}\exp(a_{c'}(x))}
\end{align}
   where $a_{c'}(x)$ denotes the activation in the $c'$th feature channel at the pixel position $x \in \Omega$ with $ \Omega \subset$ $\mathbb{Z}^{2}$, $C$ is the number of land cover categories, $p_{c}(x)$ is the approximated maximum function, which represents the probability of each pixel for each category. The cross entropy calculates the deviation of $p_{\iota(x)}(x)$ at each position.

\begin{align}
E = \sum_{x \in \Omega}\log(p_{\iota(x)}(x))
\end{align}
where $\iota : \Omega \rightarrow \{1,...,C\}$ is the ground truth of each pixel, $\iota(x)$ is the label in the location of x.

\section{Experiments}
\label{expe}
In this section, four PolSAR images are described in detail, which are used to verify the performance of the proposed algorithm. The four images have significant representativeness, obtaining from two airborne systems and two cities, the details are listed in Table \ref{tabf1}-\ref{satellitetb}. The parameter settings of the proposed method are discussed. Meanwhile, evaluation metrics are given.

\begin{figure*}[htb]
  \centering
\begin{minipage}[b]{0.38\linewidth}
  \centering
\centerline{\hspace{0.34cm}\epsfig{figure= 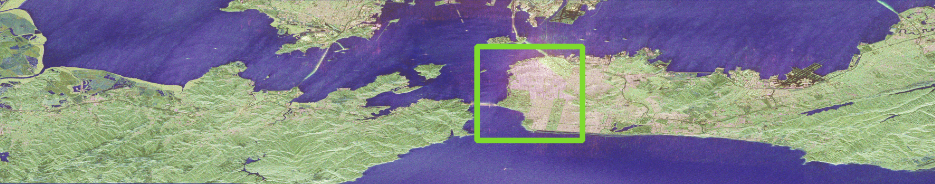,width=16.1cm}}
  \vspace{0.25cm}
\end{minipage}
\caption{Pauli RGB image of the San Francisco big figure. The whole data of this image downloaded from the RADARSAT-2 web \cite{dataset1}, and the area in box is often used. The area coordinate is (7326:9125,661:2040).}
\label{sanbig}
\end{figure*}

\begin{figure*}[htb]
  \centering
\begin{minipage}[b]{0.38\linewidth}
  \centering
\centerline{\hspace{0.34cm}\epsfig{figure= 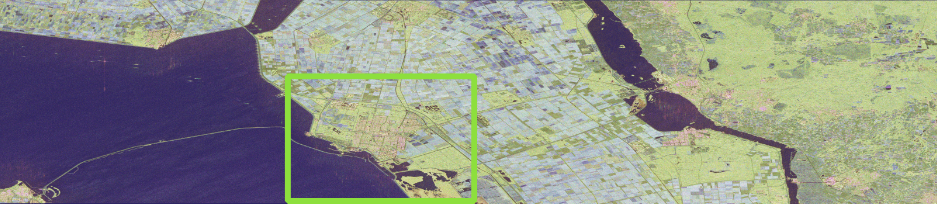,width=16.1cm}}
  \vspace{0.25cm}
\end{minipage}
\caption{Pauli RGB image of the Flevoland big figure. The whole data of this image downloaded from the RADARSAT-2 web \cite{dataset1}, and the area in box is often used. The area coordinate is (4061:6435,97:1731).}
\label{fevbig}
\end{figure*}

\subsection{Data set description}
For our experiments and evaluations, we select four PolSAR
images from a spaceborne system (Canadian Space Agency RADARSAT-2) and  an airborne system (NASA/JPL-Caltech AIRSAR). The AIRSAR supports full polarimetric modes for C, L, and P-bands where we focus on the L and C-bands. RADARSAT-2 works in C-band with the support of full polarimetric mode. The four selected
PolSAR pseudo images are from two different areas, including Flevoland, Netherlands, and the San Francisco bay area, California, USA.
In the four PolSAR images, the first two images come from Canadian Space Agency RADARSAT-2 \cite{dataset1}, as shown in the Fig. \ref{sanbig} and Fig. \ref{fevbig}. The latter two are from NASA/JPL-Caltech AIRSAR\cite{dataset2}.
We consider that this setup can fully test the performance of the algorithm over a variety of PolSAR images in terms of the system (AIRSAR and RADARSAT-2), the underlying classification problem and the operative band (C and L). The information about the four images is shown below.

\emph{1) San Francisco, RADARSAT-2, C-Band}

The area around the bay of San Francisco with the golden gate bridge is probably one of the most used scenes in PolSAR image classification over the past decades. It provides a good coverage of both natural (e.g., water, vegetation) and man-made targets
(e.g., high-density, low-density and developed).

\begin{figure}[!htbp]
\begin{minipage}[b]{.93\linewidth}
  \centering
 \centerline{\epsfig{figure= 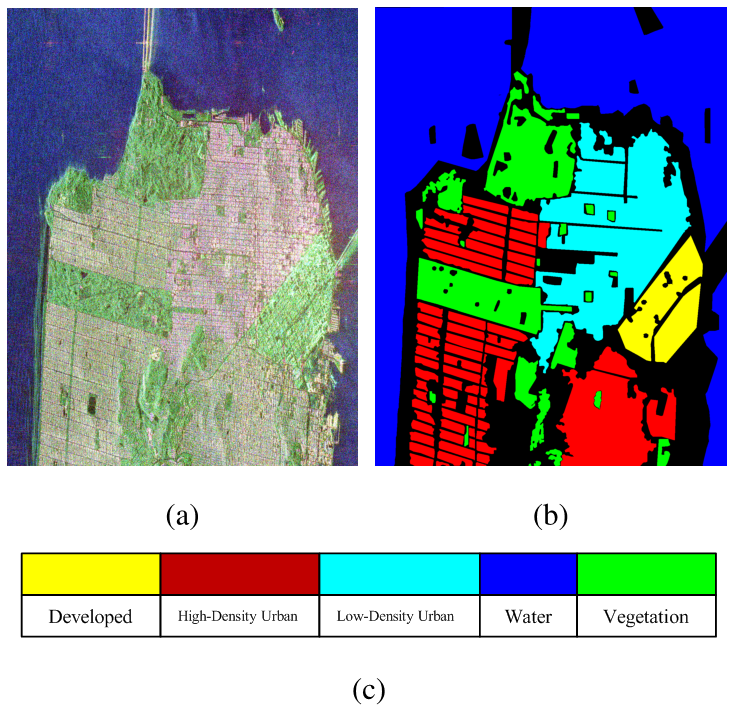, width=8.0cm}}
\end{minipage}
\caption{San Francisco pseudo image and ground truth, Radarsat-2. (a) SanFrancisco image. (b) Ground truth image. (c) The color code. }
\label{fs1}
\end{figure}

This RADARSAT-2 fully PolSAR image at fine quad-pol mode (8-m spatial resolution) was taken in April 2, 2008. The selected scene is an 1380$\times$1800-pixel subregion. The Pauli-coded pseudo color image, the used ground truth data and the color code are shown in Fig. \ref{fs1}. In the ground truth map Fig. \ref{fs1}(b), there are five kinds of objects including developed, high-density, low-density, water and vegetation, which can be simply written as c1-c5. The size of original image is 2820 $\times$ 14416, which is a relatively large scene with great research significance. Sub area coordinate is (7326:9125,661:2040), which was shown in Fig. \ref{sanbig}. This coordinate can help researchers easily find the area and know the source of the data. Table \ref{tabf1} shows the numbers of the train and test samples.

 \emph{2) Flevoland, Radarsat-2, C-Band}

\begin{figure}[!htbp]
\begin{minipage}[b]{0.93\linewidth}
  \centering
 \centerline{\epsfig{figure= 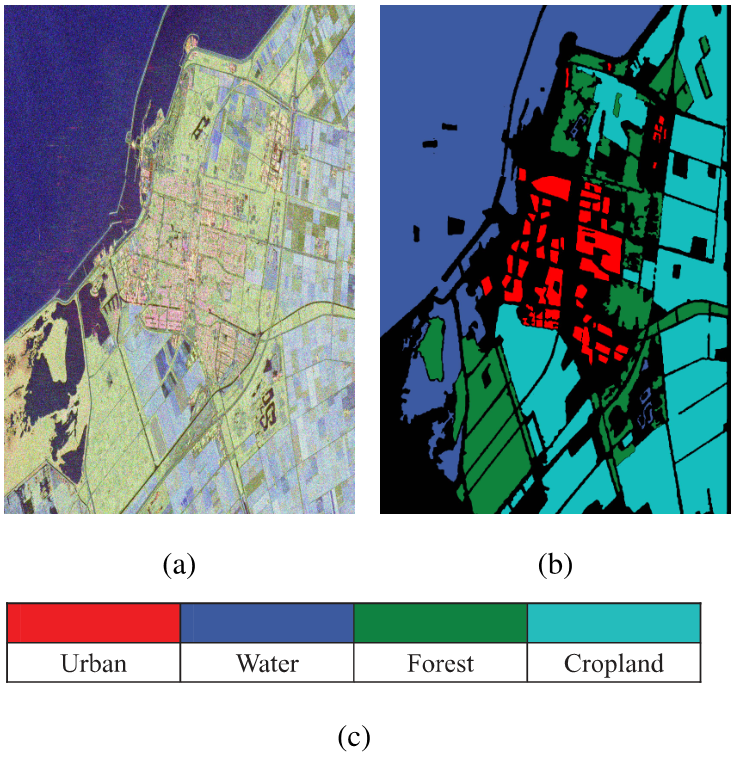,width=8.0cm}}
\end{minipage}
\caption{Flevoland pseudo image and ground truth, Radarsat-2. (a) Flevoland image. (b) Ground truth image. (c) The color code. }
\label{fs2}
\end{figure}

This RADARSAT-2 fully PolSAR image at one quad-pol mode (8-m spatial resolution) of Flevoland, the Netherlands, was taken in April 2, 2008.
The selected scene is a 1635 $\times$ 2375-pixel subregion, which mainly contains four terrain classes: 1) woodland/forest;
2) cropland; 3) water; and 4) urban area. The Pauli color coded image, the ground truth data and the color code are shown in Fig. \ref{fs2}.
The size of original image is 2820 $\times$ 12944.
Sub area coordinate is (4061:6435,97:1731), as was shown in Fig. \ref{fevbig}.  Table \ref{tabf2} shows the number of the train and test samples.

\emph{3) San Francisco, AIRSAR, L-Band}

\begin{figure}[!htbp]
\begin{minipage}[b]{1\linewidth}
  \centering
 \centerline{\epsfig{figure= 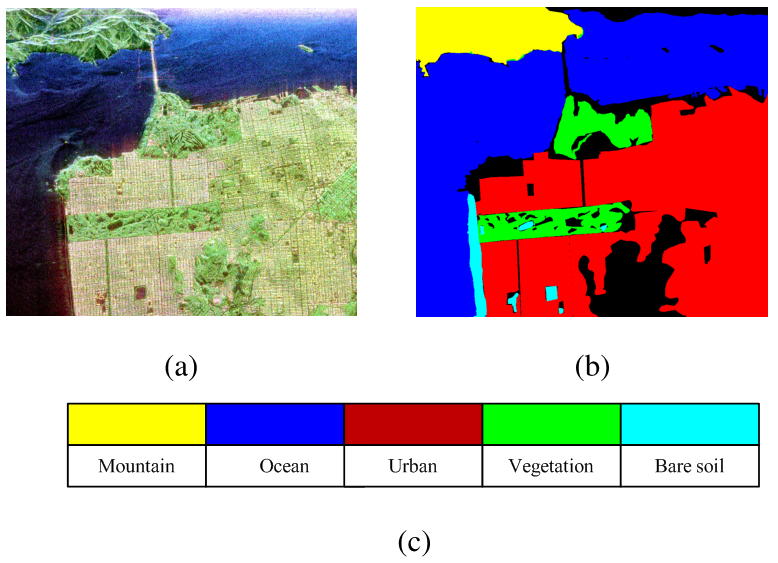,width=8.0cm}}
\end{minipage}
\caption{Sanfrancisco pseudo image and ground truth, AIRSAR. (a) Sanfrancisco image. (b) Ground truth image. (c) The color code. }
\label{fs3}
\end{figure}

This PolSAR image of San Francisco bay has been used in many literature works, Fig. \ref{fs3} show the Pauli RGB image, the ground truth map and the color code. The size of this image is 900$\times$1024. The spatial resolution is 10 m for 20 MHz.
Pixels in this image can be classified into five categories, abbreviated letter c1-c5 indicate the categories of mountain, ocean, urban, vegetation and bare soil, respectively. Table \ref{tabf3} shows the number of the train and test samples.

\emph{4) Flevoland, AIRSAR, L-Band}

\begin{figure}[!htbp]
\begin{minipage}[b]{1\linewidth}
  \centering
 \centerline{\epsfig{figure= 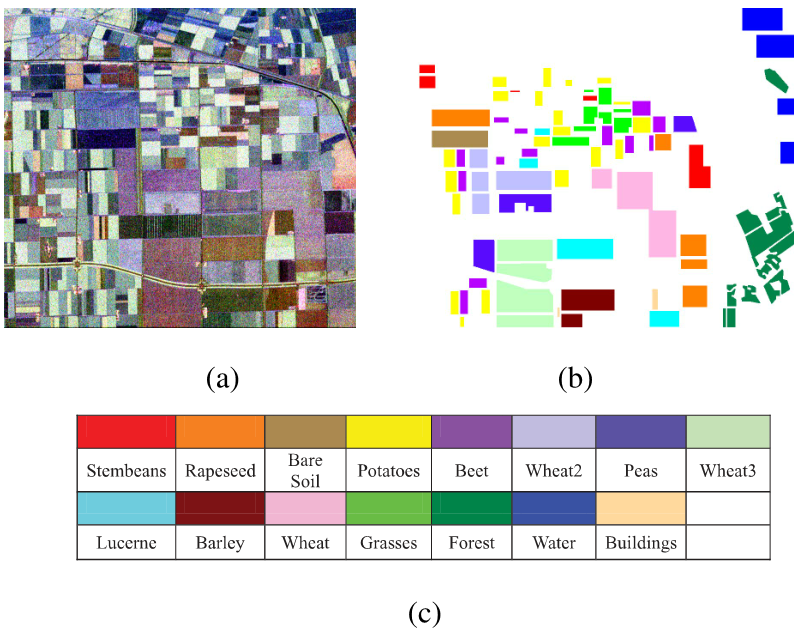,width=8.0cm}}
\end{minipage}
\caption{Flevoland pseudo image and ground truth, AIRSAR. (a) Flevoland image. (b) Ground truth image. (c) The color code. }
\label{fs4}
\end{figure}

The PolSAR image of Flevoland is shown in Fig. \ref{fs4}(a), there are 15 categories in the ground truth map Fig. \ref{fs4}(b), and the color code is shown in Fig. \ref{fs4}(c). It is the picture with the most kinds of objects in the public PolSAR data collection at present.
The spatial resolution is 10 m for 20 MHz. The size of this PolSAR data is 750 $\times$ 1024.
There are 15 kinds of objects to be identified, including stem beans, rapeseed, bare soil, potatoes, beet, wheat2, peas, wheat3, lucerne, barley, wheat, grasses, forest, water and building. These objects are simply written as c1-c15 in this paper. The numbers of the train and test samples are shown in Table \ref{tabf4}.
\begin{table}[htbp]
\newcommand{\tabincell}[2]{\begin{tabular}{@{}#1@{}}#2\end{tabular}}
\centering
\scriptsize
\caption{ \protect\ Land classes and numbers of pixels in the first data set.}
\begin{tabular}{c|c|c|c}
\hline
\hline
class code & name &\tabincell{c}{No. of training \\ samples} & \tabincell{c}{No. of testing \\ samples}\\
\hline
1 & Water & 1000 & 852078\\
2 & Vegetation & 1000 & 237237\\
3 & High-Density Urban & 1000 & 351181\\
4 & Low-Density Urban  & 1000 & 282975\\
5 & Developed & 1000 & 80616\\
\hline
\hline
\end{tabular}
  \label{tabf1}
\end{table}
\begin{table}[htbp]
\newcommand{\tabincell}[2]{\begin{tabular}{@{}#1@{}}#2\end{tabular}}
\centering
\scriptsize
\caption{ \protect\ Land classes and numbers of pixels in the second data set.}
\begin{tabular}{c|c|c|c}
\hline
\hline
class code & name &\tabincell{c}{No. of training \\ samples} & \tabincell{c}{No. of testing \\ samples}\\
\hline
1 & Urban & 1000 & 136561\\
2 & Water & 1000 & 1050726\\
3 & Forest & 1000 & 403584\\
4 & Cropland & 1000 & 935821\\
\hline
\hline
\end{tabular}
  \label{tabf2}
\end{table}

\begin{table}[htbp]
\newcommand{\tabincell}[2]{\begin{tabular}{@{}#1@{}}#2\end{tabular}}
\centering
\scriptsize
\caption{ \protect\ Land classes and numbers of pixels in the third data set.}
\begin{tabular}{c|c|c|c}
\hline
\hline
class code & name &\tabincell{c}{No. of training \\ samples} & \tabincell{c}{No. of testing \\ samples}\\
\hline
1 & Mountain & 1000 & 13701\\
2 & Ocean & 1000 & 62731\\
3 & Urban & 1000 & 329566\\
4 & Vegetation & 1000 & 342795\\
5 & Bare soil & 1000 & 53509\\
\hline
\hline
\end{tabular}
  \label{tabf3}
\end{table}

\begin{table}[htbp]
\newcommand{\tabincell}[2]{\begin{tabular}{@{}#1@{}}#2\end{tabular}}
\centering
\scriptsize
\caption{ \protect\ Land classes and numbers of pixels in the fourth data set.}
\begin{tabular}{c|c|c|c}
\hline
\hline
class code & name &\tabincell{c}{No. of training \\ samples} & \tabincell{c}{No. of testing \\ samples}\\
\hline
1 & Water & 1000 & 12232\\
2 & Barely & 1000 & 6595\\
3 & Peas & 1000 & 8582\\
4 & Stem beans & 1000 & 5338\\
5 & Beet & 1000 & 9033\\
6 & Forest & 1000 & 17044\\
7 & Bare soil & 1000 & 4109\\
8 & Grasses & 1000 & 6058\\
9 & Rapeseed & 1000 & 12863\\
10 & Lucerne & 1000 & 9181\\
11 & Wheat2 & 1000 & 10159\\
12 & Wheat1 & 1000 & 15386\\
13 & Buildings & 200 & 535\\
14 & Potatoes & 1000 & 15156\\
15 & Wheat3 & 1000 & 21241\\
\hline
\hline
\end{tabular}
  \label{tabf4}
\end{table}

\begin{table}
\addtolength{\tabcolsep}{-0.6pt}
\tiny
\centering
\caption{Overview of polarimetric image data used within experiments}
\begin{tabular}{c|c|c|c|c|c|c}
\hline
\hline
Name & System+Band  & Abbr. & Date & Spatial Resolution & Dimensions  & class \\
\hline
San Francisco & RADARSAT-2 C & SF-RL & Apr 2008 & 8  (m) & 1380 $\times$ 1800 & 5\\
\hline
Flevoland & RADARSAT-2 C & Flevo-RC & Apr 2008 & 8  (m) & 1635 $\times$ 2375 & 4 \\
\hline
San Francisco & AIRSAR L & SF-AL & Aug 1989 & 10 (m) & 1024 $\times$ 900 & 5 \\
\hline
Flevoland & AIRSAR L & Flevo-AL & Aug 1989 & 10 (m) &1024 $\times$ 750 & 15 \\
\hline
\hline
\end{tabular}
\label{satellitetb}
\end{table}

\subsection{Experimental setup}

Traditional polarimetric features are extracted to compare with the polarimetric scattering coding in the proposed method and contrast algorithms. {In this paper, we used a 22-dimensional feature vector as the traditional polarimetric feature, including the upper right element's absolute value of the 3$\times$3 polarimetric coherency matrix $\textbf{T}$, the upper right element's absolute value of the 3$\times$3 polarimetric covariance matrix $\textbf{C}$, three component of Pauli decomposition, three components of Freeman decomposition \cite{Freeman1993Three},and four components of Yamaguchi decomposition \cite{Yamaguchi2011Four}, expressed as PF22.} The feature extracted by polarimetric scattering coding can be written as SSCF. In the experiment, we consider two aspects for comparison experiments: feature extraction and classifier design. For feature extraction, the commonly feature PF22 and the proposed SSCF are adopted. For classifier design, the traditional algorithms maximum likelihood (MLD) \cite{strahler1980use} and support vector machine (SVM) \cite{hearst1998support} are used to classify the image data, the deep learning algorithms also be adopted, such as convolutional neural network (CNN) \cite{Krizhevsky2012ImageNet}, PFDCN \cite{Chen2018PolSAR} and full convolutional network (FCN) \cite{long2015fully}. Based on these considerations, the comparison methods include PF22-MLD, PF22-SVM, PF22-CNN, SSC-CNN, PF22-FCN and PFDCN. The proposed method can be represented as PCN.

In contrast algorithms, in order to make as fair a comparison as possible, The standard deviation has been added in the result through ten random experiments, the parameters of CNN and FCN are set to the same as possible.
The CNN is structured as follows, the first convolutional layer filters the input image patch with 64 kernels of size 5$\times$5. The second convolutional layer takes the output of the first convolutional layer as the input and filers it with 128 kernels of size 5$\times$5.
The third, fourth, and fifth convolutional layers are connected to one another without any intervening
pooling or normalization layers. The third convolutional layer has 256 kernels of size 5$\times$5 connected to the (normalized, pooled) outputs of the second convolutional layer. The fourth convolutional layer has 256 kernels of size 5$\times$5 , and the fifth convolutional layer has 128 kernels of size 5$\times$5. At last, The fully-connected layer has 100 neurons.

The structure of FCN is as follows, the relevant setting of the first five layers is the same as that of the CNN, the latter four layers is corresponding to the first five layers and the original image size. Skip layer also been adopted at the FCN's sixth to eighth layers to keep image edge information. In Fig. \ref{flowchart}, the last feature maps of the two FCN are stacked. And then, the output layer then performs a 1$\times$1 convolution to produce the same number of feature maps as the number of classes in each data set. In the case that the input is polarimetric scattering coding matrix, we added a specially designed two layer networks in front of CNN and FCN, the first layer is 32 kernels of size 4$\times$4 with a stride of 1 pixel, the second layer is 64 kernels of size 4$\times$4 with a stride of 2 pixels.
Through this network, we can get the classification map with the same size as the original image.
We train the model in a single step of optimization, and the weight is initialized by $Xavier$ \cite{Xavier2010}. The stochastic gradient descent with momentum 0.9 is used to train the weights of the model. The initial learning rate is set to 0.01, the train batch size is 1.
In the experiments, we randomly selected 1000 points per class for training, and the remaining labeled samples were used for testing. In addition to training data, other samples do not participate in training and learning.
In SVM, the kernel function is the Radial Basis Function (RBF). The multiclass strategy is one-versus-rest.
In the contrast algorithm, the size of image patch is 32$\times$32. In the proposed method, the training pixel is activated, others is set to 0 in the label, which is the ignore label in the training stage.

All the experiments are running on a HP Z840 workstation with an Intel Xeon(R) CPU, a GeForce GTX TITAN X GPU, and 64G RAM under Ubuntu 16.04 LTS. All of these methods are implemented using the deep learning framework of TensorFlow.

\subsection{Evaluation metrics}
For evaluating the classification performance, the experiment results were assessed by single class recall rate, overall accuracy (OA), average accuracy (AA) and kappa coefficient (Kappa). Overall accuracy can be defined as follows,
\begin{align}
OA = \frac{M}{N}
\end{align}
where $M$ is the number of classified correctly, and $N$ is the total number of samples.
Average accuracy can also be write as follows,
\begin{align}
AA = \frac{1}{C}\sum_{i=1}^{C}\frac{M_{i}}{N_{i}}
\end{align}
where $C$ is the number of categories, $i$ category index, $M_{i}$ is the number of correct samples for the $i$-th categories, $N_{i}$ is the number of samples in the $i$-th class. $\frac{M_{i}}{N_{i}}$ is the $i$-th class recall rate. The formula of kappa coefficient can be written as,
\begin{align}
Kappa = \frac{OA-P}{1-P}, P = \frac{1}{N^{2}}\sum_{i=1}^{C}\bar{Z}(i,:)*\bar{Z}(:,i)
\end{align}
where $OA$ is the overall accuracy, $Z$ is the confusion matrix, $\bar{Z}(i,:)$ is the sum of the  $i$-th row elements, and $\bar{Z}(:,i)$ is the sum of the  $i$-th column elements. $N$ is the total number of samples.
\section{Result and Discussions}
\begin{figure*}[!htb]
  \centering
\begin{minipage}[b]{0.30\linewidth}
 \centerline{\epsfig{figure= 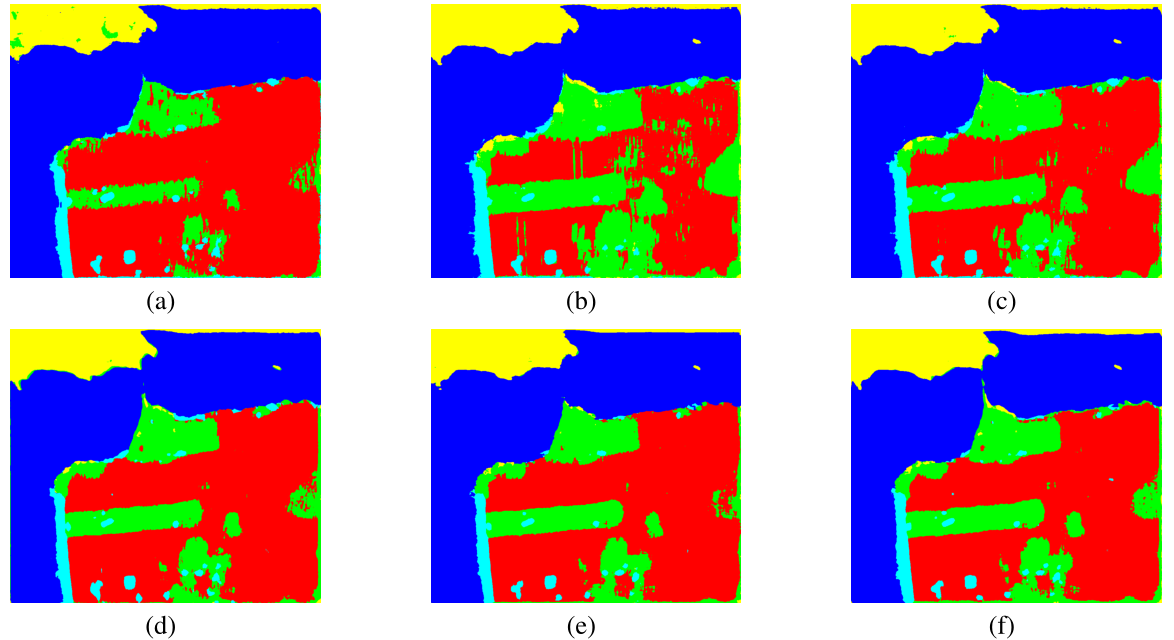,width=14cm}}
\end{minipage}
\caption{The classification maps of different methods on the first scene image. (a)-(f) : PF18-MLD, PF18-SVM, PF18-CNN, SSC-CNN, PF18-FCN and PCN }
\label{fs3rs}
\end{figure*}

\begin{figure*}[!htb]
  \centering
\begin{minipage}[b]{0.30\linewidth}
 \centerline{\epsfig{figure= 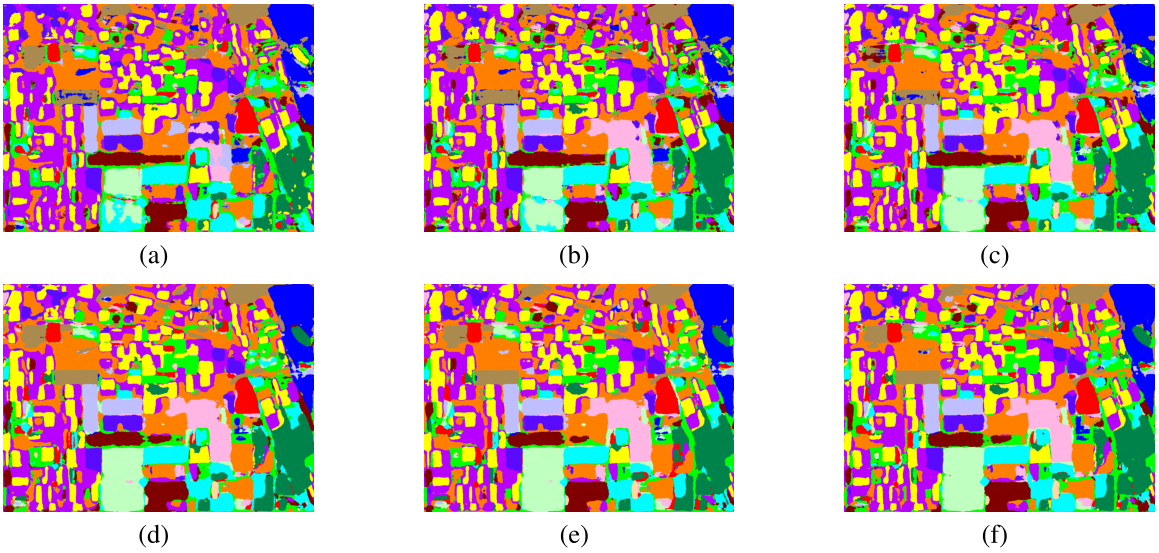,width=14cm}}
\end{minipage}
\caption{The classification maps of different methods on the first scene image. (a)-(f) : PF18-MLD, PF18-SVM, PF18-CNN, SSC-CNN, PF18-FCN and PCN }
\label{fs4rs}
\end{figure*}

\begin{figure}[!htb]
  \centering
\begin{minipage}[b]{0.48\linewidth}
 \centerline{\epsfig{figure= 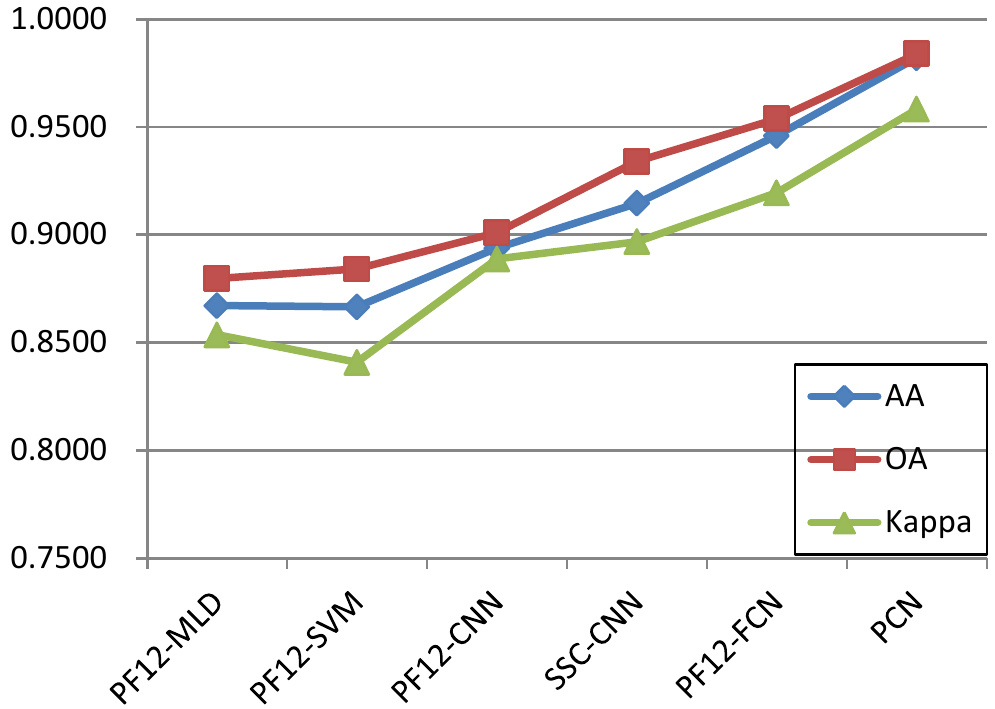,width=4.3cm}}
  \centerline{(a) }\medskip
\end{minipage}
\begin{minipage}[b]{0.48\linewidth}
 \centerline{\epsfig{figure= 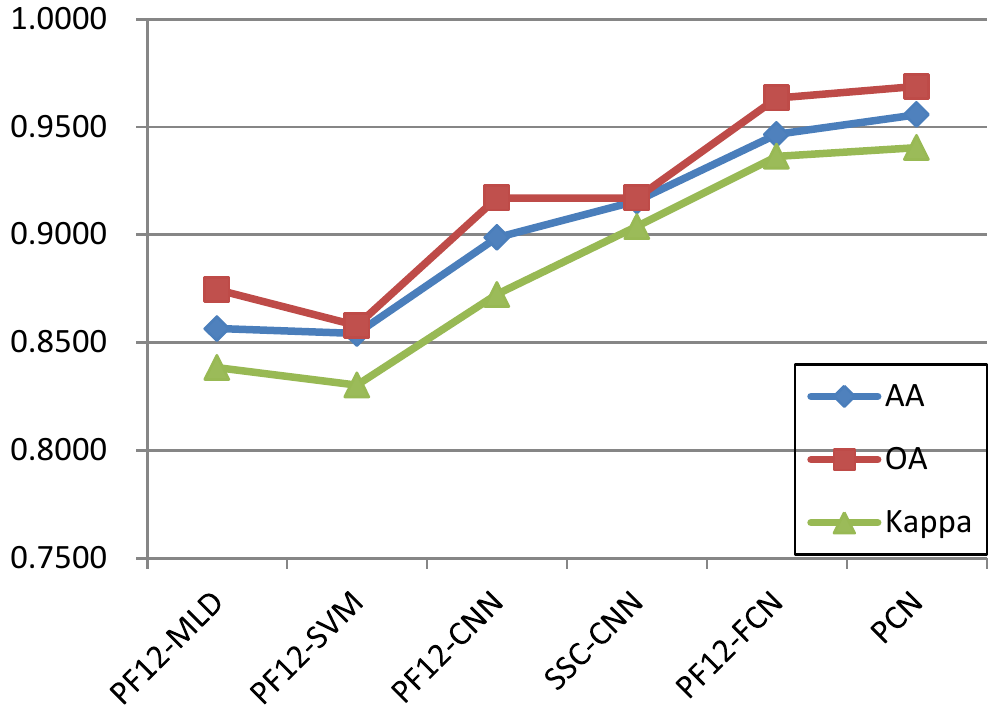,width=4.3cm}}
  \centerline{(b) }\medskip
\end{minipage}
\caption{Classification accuracies on Fig. \ref{fs3} and Fig. \ref{fs4} by different methods. (a) Classification accuracies on Fig. \ref{fs3}. (b) Classification accuracies on Fig. \ref{fs4}.}
\label{ft2}
\end{figure}

Our experimentation will be separated into four parts. In the first part, we demonstrate the proposed algorithm using the two data sets from the RADARSAT-2. In the second part, we evaluate the proposed algorithm using the other two data sets from the AIRSAR. In the third part, we give a the significance analysis of the results. Finally, we show the computation times of the proposed algorithm and comparison algorithms in part four.
The results are shown in the Fig. \ref{fs1rs} - \ref{ft2} and TABLE \ref{fs1tb} - \ref{fs4tb}.
\subsection{Experimental with the dataset from the RADARSAT-2}
\begin{figure*}[!htb]
  \centering
\begin{minipage}[b]{0.30\linewidth}
 \centerline{\epsfig{figure= 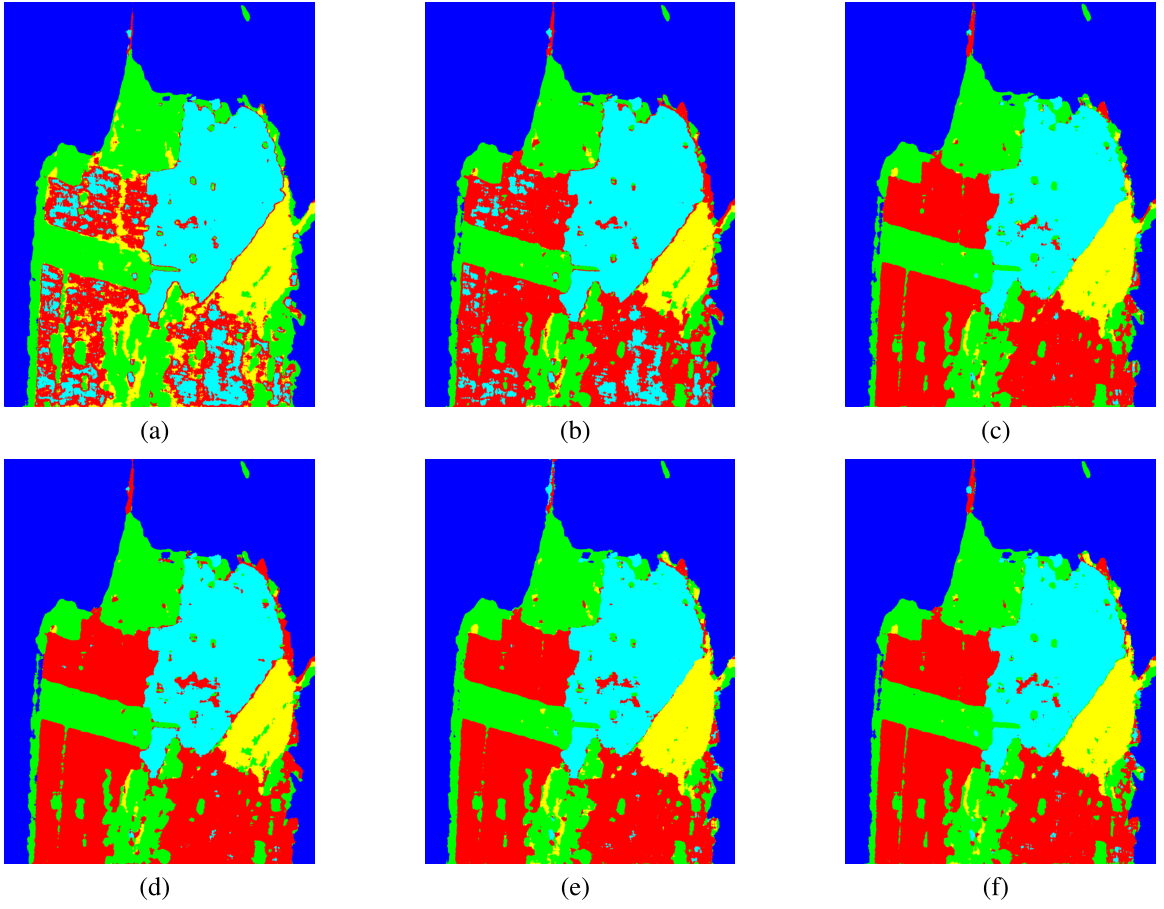,width=14cm}}
\end{minipage}
\caption{The classification maps of different methods on the first scene image. (a)-(f) : PF18-MLD, PF18-SVM, PF18-CNN, SSC-CNN, PF18-FCN and PCN }
\label{fs1rs}
\end{figure*}

\begin{figure*}[!htb]
  \centering
\begin{minipage}[b]{0.30\linewidth}
 \centerline{\epsfig{figure= 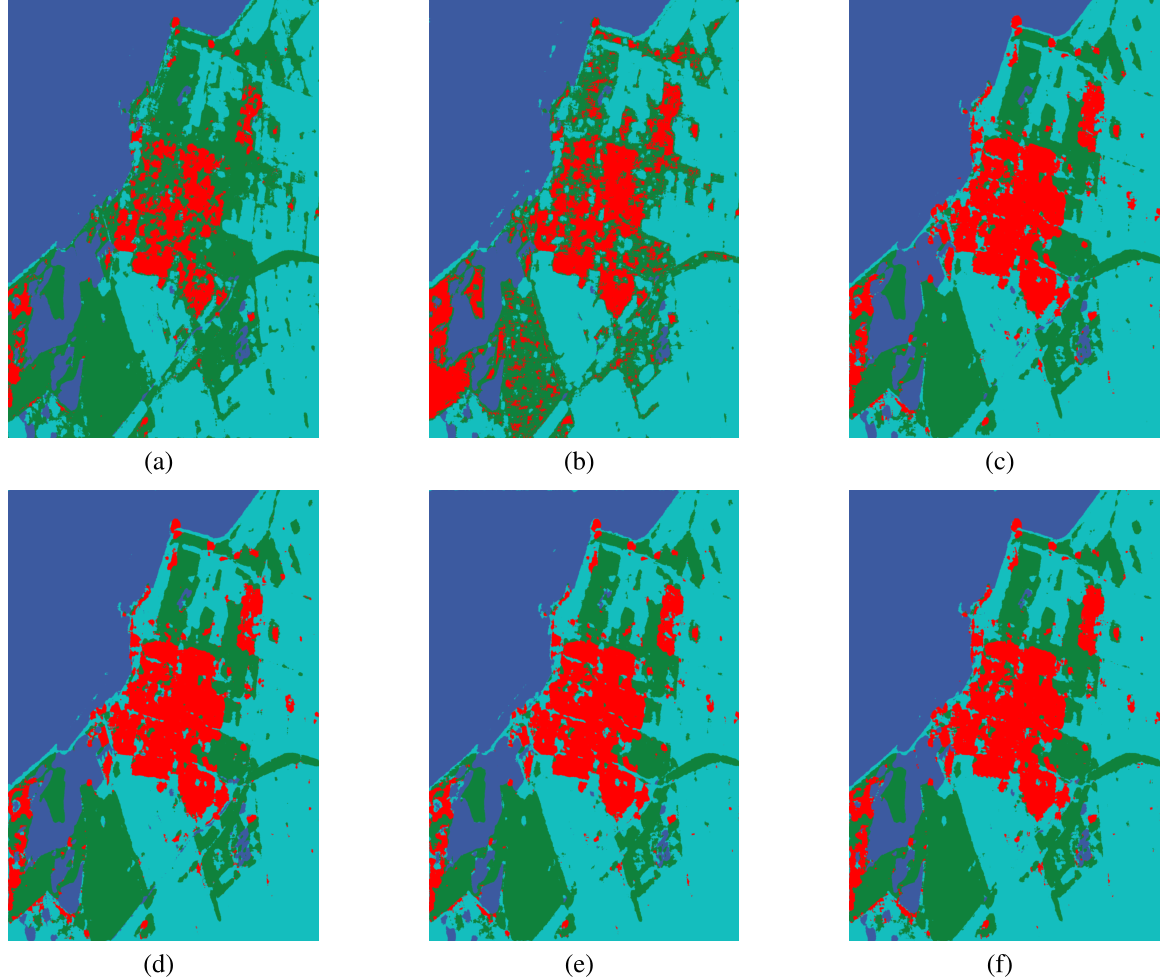,width=14cm}}
\end{minipage}
\caption{The classification maps of different methods on the first scene image. (a)-(f) : PF18-MLD, PF18-SVM, PF18-CNN, SSC-CNN, PF18-FCN and PCN }
\label{fs2rs}
\end{figure*}

\begin{figure}[!htb]
  \centering
\begin{minipage}[b]{0.48\linewidth}
 \centerline{\epsfig{figure= 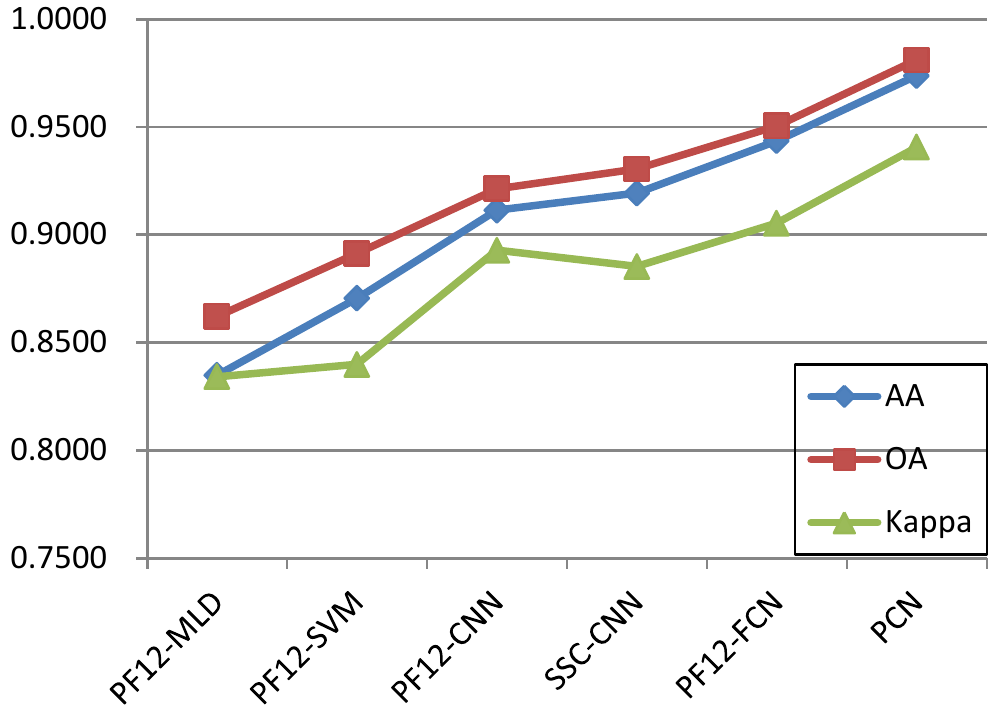,width=4.3cm}}
  \centerline{(a) }\medskip
\end{minipage}
\begin{minipage}[b]{0.48\linewidth}
 \centerline{\epsfig{figure= 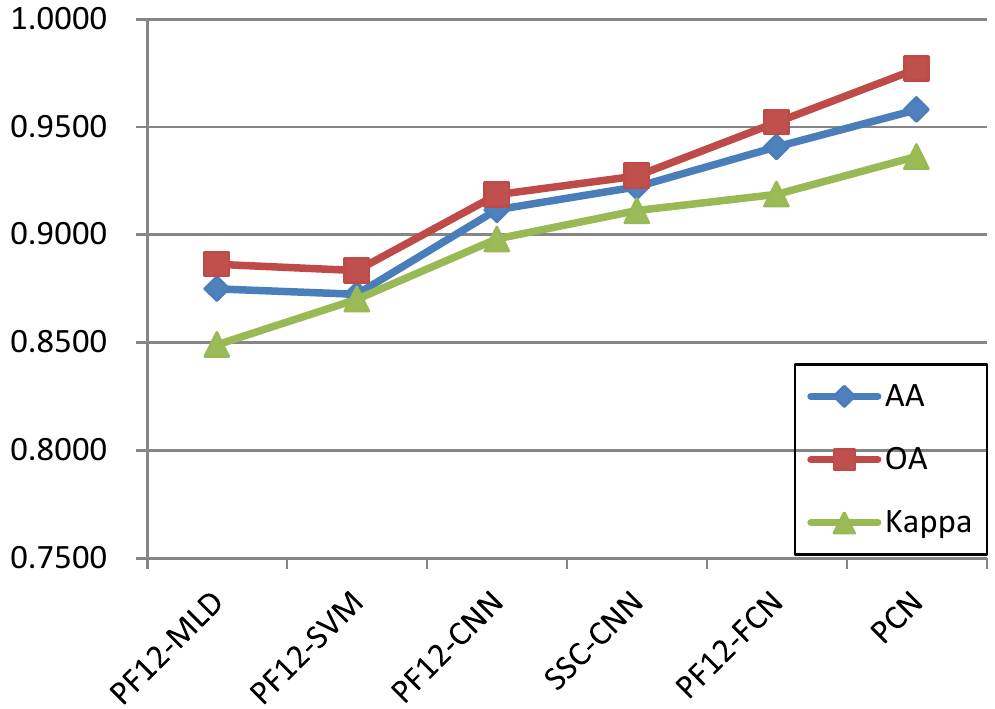,width=4.3cm}}
  \centerline{(b) }\medskip
\end{minipage}
\caption{Classification accuracies on Fig. \ref{fs1} and Fig. \ref{fs2} by different methods. (a) Classification accuracies on Fig. \ref{fs1}. (b) Classification accuracies on Fig. \ref{fs2}.}
\label{ft1}
\end{figure}

The classification results from different algorithms are demonstrated in Fig. \ref{fs1rs}(a)-(g) and Fig. \ref{fs2rs}(a)-(g),
and the accuracies for each class are listed in Table \ref{fs1tb} and Table \ref{fs2tb}, respectively.
Fig. \ref{ft1}(a)-(b) show a clear contrast of classification accuracy, the trend line is generated by the data in Table \ref{fs1tb} and Table \ref{fs2tb}.
It can be seen that the performance of the proposed method is better than others. The classification accuracies are higher than the contrast algorithms. The classification maps are closer to the ground truth. Trend chart of accuracy in Fig. \ref{ft1}(a)-(b) show that the PCN is better than others.

Fig. \ref{fs1rs} shows that PF22-MLD and PF22-SVM cannot distinguish high-density urban well, and mistake much high-density urban for low-density urban, which is due to the fact that the two objects are similar, and the method is difficult to distinguish its characteristics.
The result of FCN is better than the CNN's, which can be seen from the contiguous area of two class, such as coastal area.
The classification map of PF22-CNN is worse than that of SSC-CNN in recognizing mixed terrain.
Meanwhile, the result of PCN is better than that of PF22-FCN and PFDCN, the proposed algorithm PFDCN and PCN almost detects all terrain. Especially, the low-density urban misclassified in high-density urban is almost completely corrected.
Table \ref{fs1tb} shows that the accuracy of deep learning methods are higher than the conventional methods, with a difference of two to six percentage points. It can be seen clearly that high-density urban is misclassified. The classification accuracy of high-density urban is 80.54$\%$.

In Fig. \ref{fs2rs}, there are only four types of objects that need to be distinguished, namely urban, water, forest and cropland. Water is the most easily recognizable relatively, inland water can also be judged right. It is difficult to judge fake forest in urban and cropland.
Some urban and cropland are mistakenly divided into forests in the classification maps obtained by PF22-MLD and PF22-SVM. The classification methods based on convolution get the better classification maps, some homogeneous areas can be correctly obtained. Further more, the performance of the proposed method is more outstanding. The urban and cropland can be seen completely, there is almost no noise.
Meanwhile, Table \ref{fs2tb} gives a quantitative result, PF22-MLD and PF22-SVM get a low accuracy, especially in urban and cropland areas, other methods show an upward trend. PCN has the highest accuracy.

\begin{table}[!htp]
\tiny
\centering
\caption{Classification accuracy (\%) of the first scene image}
\setlength{\tabcolsep}{0.7mm}{
\begin{tabular}{c|c|c|c|c|c|c|c|c}
\hline
\hline
Method & c1 & c2 & c3 & c4 & c5 & AA & OA & Kappa\\
\hline
PF22-MLD & 86.90$\pm$0.16 & 82.83$\pm$0.14 & 80.54$\pm$0.16 & 82.98$\pm$0.09 & 84.24$\pm$0.14 & 83.49$\pm$0.17 & 86.23$\pm$0.15 & 83.42$\pm$0.13 \\
\hline
PF22-SVM & 89.14$\pm$0.18 & 86.48$\pm$0.16 & 88.24$\pm$0.16 & 89.1$\pm$0.12 & 82.31$\pm$0.09 & 87.05$\pm$0.1 & 89.13$\pm$0.1 & 83.99$\pm$0.14 \\
\hline
PF22-CNN & 92.82$\pm$0.15 & 92.93$\pm$0.15 & 93.86$\pm$0.16 & 94.13$\pm$0.13 & 85.93$\pm$0.16 & 91.93$\pm$0.11 & 93.05$\pm$0.07 & 88.54$\pm$0.14 \\
\hline
SSC-CNN & 86.27$\pm$0.09 & 94.07$\pm$0.09 & 94.56$\pm$0.08 & 93.54$\pm$0.15 & 87.24$\pm$0.12 & 91.14$\pm$0.12 & 92.14$\pm$0.08 & 89.29$\pm$0.08 \\
\hline
PF22-FCN  & 93.46$\pm$0.06 & 97.39$\pm$0.13 & 92.07$\pm$0.09 & 96.43$\pm$0.14 & 92.42$\pm$0.12 & 94.35$\pm$0.11 & 95.05$\pm$0.14 & 90.54$\pm$0.09 \\
\hline
PFDCN  & 94.68$\pm$0.12 & 97.64$\pm$0.07 & 92.54$\pm$0.06 & 96.92$\pm$0.09 & 92.82$\pm$0.1 & 95.88$\pm$0.08 & 96.05$\pm$0.12 & 93.54$\pm$0.15 \\
\hline
PCN  & 96.2$\pm$0.15 & 94.83$\pm$0.15 & 98.56$\pm$0.14 & 98.5$\pm$0.12 & 98.79$\pm$0.08 & \textbf{97.44}$\pm$0.14 & \textbf{98.24}$\pm$0.07 & \textbf{95.27}$\pm$0.1 \\
\hline
\hline
\end{tabular}}
\label{fs1tb}
\end{table}


\begin{table}[!htp]
\tiny
\centering
\caption{Classification accuracy (\%) of the second scene image}
\setlength{\tabcolsep}{1.25mm}{
\begin{tabular}{c|c|c|c|c|c|c|c}
\hline
\hline
Method & c1 & c2 & c3 & c4 & AA & OA & Kappa\\
\hline
PF22-MLD & 88.63$\pm$0.07 & 81.91$\pm$0.16 & 89.90$\pm$0.18 & 86.46$\pm$0.14 & 86.71$\pm$0.12 & 87.98$\pm$0.06 & 85.36$\pm$0.08 \\
\hline
PF22-SVM & 86.47$\pm$0.09 & 86.4$\pm$0.12 & 85.16$\pm$0.08 & 88.62$\pm$0.17 & 86.66$\pm$0.16 & 88.42$\pm$0.08 & 84.09$\pm$0.07 \\
\hline
PF22-CNN & 90.95$\pm$0.17 & 85.12$\pm$0.06 & 94.39$\pm$0.06 & 87.05$\pm$0.13 & 89.38$\pm$0.15 & 90.09$\pm$0.09 & 88.88$\pm$0.08 \\
\hline
SSCF+CNN & 89.65$\pm$0.12 & 91.95$\pm$0.09 & 92.79$\pm$0.12 & 91.45$\pm$0.16 & 91.45$\pm$0.08 & 93.4$\pm$0.08 & 89.68$\pm$0.08 \\
\hline
PF22-FCN  & 91.14$\pm$0.10 & 95.64$\pm$0.08 & 93.12$\pm$0.11 & 98.52$\pm$0.11 & 94.6$\pm$0.08 & 95.38$\pm$0.09 & 91.95$\pm$0.07 \\
\hline
PFDCN  & 94.33$\pm$0.10 & 96.08$\pm$0.16 & 94.16$\pm$0.15 & 97.66$\pm$0.13 & 95.6$\pm$0.13 & 96.21$\pm$0.14 & 93.54$\pm$0.06 \\
\hline
PCN  & 98.86$\pm$0.11 & 98.9$\pm$0.13 & 99.31$\pm$0.13 & 95.76$\pm$0.08 & \textbf{98.23$\pm$0.08} & \textbf{98.41$\pm$0.08} & \textbf{95.84$\pm$0.13} \\
\hline
\hline
\end{tabular}}
\label{fs2tb}
\end{table}
\begin{table}[!htp]
\tiny
\centering
\caption{Classification accuracy (\%) of the third scene image}
\setlength{\tabcolsep}{0.7mm}{
\begin{tabular}{c|c|c|c|c|c|c|c|c}
\hline
\hline
Method & c1 & c2 & c3 & c4 & c5 & AA & OA & Kappa\\

\hline
PF22-MLD & 88.27$\pm$0.11 & 86.83$\pm$0.22 & 85.52$\pm$0.18 & 89.93$\pm$0.13 & 86.97$\pm$0.06 & 87.50$\pm$0.09 & 88.64$\pm$0.11 & 84.91$\pm$0.07 \\
\hline
PF22-SVM & 84.75$\pm$0.17 & 89.73$\pm$0.11 & 87.94$\pm$0.15 & 82.79$\pm$0.14 & 90.98$\pm$0.16 & 87.24$\pm$0.11 & 88.35$\pm$0.14 & 87.01$\pm$0.09 \\
\hline
PF22-CNN & 91.32$\pm$0.12 & 92.10$\pm$0.11 & 91.74$\pm$0.08 & 86.56$\pm$0.15 & 94.08$\pm$0.12 & 91.16$\pm$0.12 & 91.87$\pm$0.13 & 89.81$\pm$0.10 \\
\hline
SSC-CNN & 91.27$\pm$0.15 & 94.01$\pm$0.15 & 90.90$\pm$0.09 & 93.37$\pm$0.12 & 91.59$\pm$0.14 & 92.22$\pm$0.07 & 92.74$\pm$0.06 & 91.12$\pm$0.06 \\
\hline
PF22-FCN  & 96.45$\pm$0.07 & 94.39$\pm$0.06 & 91.22$\pm$0.13 & 94.28$\pm$0.16 & 94.09$\pm$0.08 & 94.08$\pm$0.06 & 95.22$\pm$0.21 & 91.89$\pm$0.08 \\
\hline
PFDCN  & 97.08$\pm$0.09 & 95.04$\pm$0.10 & 91.46$\pm$0.09 & 95.21$\pm$0.11 & 94.89$\pm$0.14 & 94.88$\pm$0.08 & 96.21$\pm$0.09 & 92.34$\pm$0.08 \\
\hline
PCN  & 93.54$\pm$0.07 & 94.23$\pm$0.08 & 97.19$\pm$0.09 & 97.59$\pm$0.11 & 96.45$\pm$0.10 & \textbf{95.82$\pm$0.10} & \textbf{97.73$\pm$0.08} & \textbf{94.65$\pm$0.07} \\
\hline
\hline
\end{tabular}}
\label{fs3tb}
\end{table}
\begin{table*}[!htp]
\centering
\tiny
\caption{Classification accuracy (\%) of the fourth scene image}
\setlength{\tabcolsep}{0.3mm}{
\begin{tabular}{c|c|c|c|c|c|c|c|c|c|c|c|c|c|c|c|c|c|c}

\hline
\hline
Method & c1 & c2 & c3 & c4 & c5 & c6 & c7 & c8 & c9 & c10 & c11 & c12 & c13 & c14 & c15 & AA & OA & Kappa\\
\hline
PF22-MLD & 88.32$\pm$0.11 & 88.61$\pm$0.11 & 84.61$\pm$0.08 & 85.31$\pm$0.14 & 83.83$\pm$0.12 & 82.58$\pm$0.07 & 86.08$\pm$0.12 & 85.86$\pm$0.15 & 82.29$\pm$0.12 & 81.97$\pm$0.12 & 88.29$\pm$0.15 & 84.61$\pm$0.13 & 88.52$\pm$0.13 & 88.81$\pm$0.09 & 85.06$\pm$0.11 & 85.65$\pm$0.10 & 87.46$\pm$0.15 & 83.85$\pm$0.06 \\
\hline
PF22-SVM & 82.47$\pm$0.11 & 83.89$\pm$0.08 & 88.37$\pm$0.09 & 89.20$\pm$0.11 & 83.13$\pm$0.11 & 82.7$\pm$0.09 & 82.08$\pm$0.08 & 87.49$\pm$0.10 & 86.24$\pm$0.10 & 86.78$\pm$0.13 & 89.57$\pm$0.11 & 81.15$\pm$0.08 & 88.15$\pm$0.11 & 87.3$\pm$0.15 & 82.81$\pm$0.06 & 85.42$\pm$0.16 & 85.8$\pm$0.07 & 83.04$\pm$0.16 \\
\hline
PF22-CNN & 86.45$\pm$0.14 & 86.19$\pm$0.13 & 85.02$\pm$0.12 & 88.82$\pm$0.11 & 92.62$\pm$0.12 & 94.72$\pm$0.11 & 93.78$\pm$0.13 & 89.53$\pm$0.11 & 89.98$\pm$0.14 & 92.52$\pm$0.08 & 85.56$\pm$0.13 & 93.41$\pm$0.10 & 93.36$\pm$0.11 & 87.98$\pm$0.10 & 88.22$\pm$0.14 & 89.88$\pm$0.08 & 91.71$\pm$0.07 & 87.24$\pm$0.06 \\
\hline
SSC-CNN & 89.05$\pm$0.08 & 87.07$\pm$0.09 & 88.33$\pm$0.07 & 89.39$\pm$0.13 & 93.21$\pm$0.07 & 95.87$\pm$0.11 & 89.36$\pm$0.11 & 95.94$\pm$0.10 & 92.83$\pm$0.08 & 92.64$\pm$0.08 & 92.41$\pm$0.14 & 87.04$\pm$0.11 & 92.23$\pm$0.06 & 93.52$\pm$0.07 & 94.63$\pm$0.13 & 91.56$\pm$0.10 & 91.78$\pm$0.11 & 90.38$\pm$0.07 \\
\hline
PF22-FCN  & 97.92$\pm$0.13 & 93.78$\pm$0.09 & 92.78$\pm$0.14 & 95.84$\pm$0.10 & 96.25$\pm$0.07 & 92.41$\pm$0.08 & 97.71$\pm$0.08 & 93.60$\pm$0.09 & 90.25$\pm$0.12 & 90.10$\pm$0.16 & 92.19$\pm$0.14 & 93.37$\pm$0.10 & 93.76$\pm$0.12 & 91.31$\pm$0.10 & 90.66$\pm$0.12 & 94.65$\pm$0.16 & 96.34$\pm$0.11 & 93.64$\pm$0.14 \\
\hline
PFDCN  & 98.33$\pm$0.10 & 94.32$\pm$0.12 & 93.55$\pm$0.11 & 96.34$\pm$0.09 & 96.55$\pm$0.10 & 93.87$\pm$0.13 & 98.51$\pm$0.16 & 97.96$\pm$0.13 & 92.54$\pm$0.10 & 91.13$\pm$0.08 & 97.35$\pm$0.12 & 94.54$\pm$0.08 & 94.05$\pm$0.08 & 98.06$\pm$0.13 & 91.33$\pm$0.07 & 95.02$\pm$0.13 & 96.67$\pm$0.07 & 93.88$\pm$0.12 \\
\hline
PCN  & 99.50$\pm$0.15 & 94.51$\pm$0.09 & 95.29$\pm$0.10 & 96.44$\pm$0.15 & 97.59$\pm$0.10 & 95$\pm$0.11 & 96.45$\pm$0.10 & 94.33$\pm$0.08 & 93.98$\pm$0.07 & 93.51$\pm$0.16 & 95.03$\pm$0.10 & 95.2$\pm$0.10 & 95.58$\pm$0.13 & 95.89$\pm$0.08 & 95.43$\pm$0.08 & \textbf{95.59$\pm$0.11} & \textbf{96.94$\pm$0.13} & \textbf{94.08$\pm$0.07} \\
\hline
\hline
\end{tabular}}
\label{fs4tb}
\end{table*}


\subsection{Experimental with the dataset from the AIRSAR}

In this section, Fig. \ref{fs3rs}(a)-(g) and Fig. \ref{fs4rs}(a)-(g) show the classification maps from the proposed method and the contrast algorithms.
Table \ref{fs3tb} and Table \ref{fs4tb} list the statistical classification accuracy from the proposed method and the compared algorithms. A clear exhibition of classification accuracy is shown in Fig \ref{ft2}(a)-(b).

From the above results, we can get that the proposed method have the best performances. The misclassified pixels are much less in the classification maps. The classification accuracy are 95.82$\%$ and 95.59$\%$, respectively. It is 2-8 percentage points higher than other algorithms.

In Fig. \ref{fs3rs}, there are five kinds of objects to be determined, including mountain, ocean, urban, vegetation, bare soil. There is an island on the upper right of the image, called Alcatraz Island, it did not appear in Fig \ref{fs3rs}(a). On contrast, the island was distinguished as a mountain in Fig \ref{fs3rs}(b)-(e). Although the island has not been marked, it is still detected. Fig \ref{fs3rs}(g) shows a clearer outline. What's more, in the bare soil detection task, the difference of the algorithm performance can also be shown. The shape and area of the soil can be described more accurately, which can be seen in Fig \ref{fs3rs}(g).
The accuracy are listed in Table \ref{fs3tb}, the overall accuracy of PF22-MLD and PF22-SVM are only 87.50$\%$ and 87.24$\%$, other algorithms are larger than 90$\%$ and the proposed algorithm is as high as 95.82$\%$.

Fig. \ref{fs4rs} shows the classification results of the fourth data set. This data set is shown in Fig. \ref{fs4} and contains 15 kinds of objects, so many kinds of labeled data are extremely rare. Naturally, the problem will increase in difficulty. For instance, it is difficult to distinguish wheat1 wheat2 and wheat3, some wheat3 was wrongly divided into wheat1 and wheat2 in Fig. \ref{fs4rs}(a)-(c), Fig. \ref{fs4rs}(a) is the largest, Fig. \ref{fs4rs}(b) and Fig. \ref{fs4rs}(c) are relatively few. The proposed algorithm can judge almost all pixels correctly, which is shown in Fig. \ref{fs4rs}(g). Similarly, some potatoes are wrongly classified as peas, but the proposed method can solve this problem. In Table \ref{fs4tb}, the above experimental phenomena can be seen accurately through numbers, such as the classification accuracy of wheat1, wheat2 and wheat3 get a 5 percentage point increase.

We can see that the proposed approach outperforms the compared methods. It indicates that the encoded data through polarimetric scattering coding is easier to be identified and distinguished. At the same time, we can find that the modified fully convolutional network has a better classification performance than conventional convolutional network.

\subsection{Significance analysis}
In this section, we give a significance analysis about the experiment results by T-test score. The analysis results are shown in TABLE \ref{tabttest}. In the table, low values are good and significant. If the the value is less than 0.05, it means the result is significant.
It can be seen that the proposed approach has a significant advantage.

\begin{table}[htbp]
\newcommand{\tabincell}[2]{\begin{tabular}{@{}#1@{}}#2\end{tabular}}
\centering
\scriptsize
\caption{ \protect\ The T-Test scores between PCN and the compared methods.}
\begin{threeparttable}
\begin{tabular}{c|c|c|c}
\hline
\hline
Method & AA & OA & Kappa\\
\hline
PF22-MLD & 0.0028  & 0.0006  & 0.0002 \\
\hline
PF22-SVM & 0.0005  & 0.0207  & 0.0016 \\
\hline
PF22-CNN & 0.0067  & 0.0036  & 0.0011 \\
\hline
SSC-CNN & 0.0074  & 0.0002  & 0.0065 \\
\hline
PF22-FCN & 0.0316  & 0.0296  & 0.0502 \\
\hline
PFDCN & 0.0508  & 0.0423  & 0.0464 \\
\hline
\hline
\end{tabular}
\begin{tablenotes}
        \footnotesize
        \item[1] The significant value is less than 0.05.
\end{tablenotes}
 \end{threeparttable}
  \label{tabttest}
\end{table}

\subsection{Computation times}
In Table \ref{time}, we show the computation times achieved by different methods on the data sets.
Factors that affect the computation times include the size of image, the complexity of the data sets, and the methods.
The computation times then increase with the size of image and the complexity of the data sets.
When compared  Fig. \ref{fs1} and Fig. \ref{fs3}, the size of image is import factor that impacts the computational efficiency.
From \ref{fs2} and Fig. \ref{fs4}, it can easily be seen that the complexity of the data sets is an import factor for the computational efficiency. Since there are fifteen categories in  Fig. \ref{fs4}. At last, the speed of PF22-FCN and PCN is faster than others, the main reason is that this algorithm does not need sliding window to calculate the pixels.

\begin{table}[!htp]
\centering
\scriptsize
\addtolength{\tabcolsep}{-3pt}
\tiny
\caption{Computation times with different methods on data sets.}
\begin{tabular}{l|c|c|c|c|c|c|c|c}
\hline
\hline

\multirow{2}{*}{\diagbox{Methods}{Data Sets}} &\multicolumn{2}{c|}{Fig. \ref{fs1}}&\multicolumn{2}{c|}{Fig. \ref{fs2}}&\multicolumn{2}{c|}{Fig. \ref{fs3}}&\multicolumn{2}{c}{Fig. \ref{fs4}}\\
\cline{2-9}
&training(s)& testing(s) &training(s)& testing(s)&training(s)& testing(s)&training(s)& testing(s)\\
\hline
PF22-MLD & 380 & 131 & 355 & 118 &435 & 155 & 335 & 117 \\
\hline
PF22-SVM & 432 & 156 & 420 & 137& 487 & 178& 430 & 125  \\
\hline
PF22-CNN & 300 & 60 & 289 & 57& 354 & 86& 300 & 55  \\
\hline
SSC-CNN & 283 & 56 & 276 & 54& 348 & 75& 289 & 52  \\
\hline
PF22-FCN & 321 & 40 & 315 & 38& 361 & 55& 325 & 33  \\
\hline
PFDCN & 285 & 58 & 284 & 54& 351 & 76& 279 & 51  \\
\hline
PCN    & 350 & 42 & 328 & 40& 383 & 57& 341 & 35  \\
\hline
\hline
\end{tabular}
\label{time}
\end{table}
\section{Conclusions}
In this paper, we proposed a new PolSAR image classification method named polarimetric convolutional network, which based on polarimetric scattering coding and fully convolutional network. polarimetric scattering coding can keep the structure information of scattering matrix, and avoid breaking the matrix into a one-dimensional vector.

Coincidentally, convolution network needs a 2-D input, where polarimetric scattering coding matrix meets this condition.
We design an improved full convolutional network to classify data encoded by polarimetric scattering coding.
In order to make the experiment more fully and effectively, the experimental data sets consist of four data sets from two satellites, the contrast algorithms include traditional methods and latest methods.
Experimental results show that the proposed algorithm PCN is robust and get a better results, the classification maps of the proposed method are very close to the ground truth maps, and the classification accuracies are higher than contrast algorithms.
These results are mainly due to the fact that the proposed algorithm can preserve the structural semantic information of the image in the raw data.
In contrast algorithms, PF22-MLD, PF22-SVM and PF22-CNN do not give full consideration to the structural semantic information, SSC-CNN only thinks about the structural semantic information of the raw data, and PF22-FCN only considers the structural semantic information of model. The results of PFDCN outperform other comparison algorithms. Experimental results can also confirm the above inference.
From comparative experiments, we also found that this polarimetric scattering coding is effective.
For this coding, the performance of designed classification network is better.

\section*{Acknowledgment}
The authors would like to thank the NASA/JPL-Caltech and Canadian Space Agency for kindly providing the polarimetric AIRSAR/RADARSAT2 data used in this paper. The authors would also like to thank the anonymous reviewers for their helpful comments.

\ifCLASSOPTIONcaptionsoff
  \newpage
\fi



%
\bibliographystyle{IEEEtran}
\bibliography{IEEEabrv,ref}
\begin{IEEEbiography}[{\includegraphics[width=1in,height=1.25in,clip,keepaspectratio]{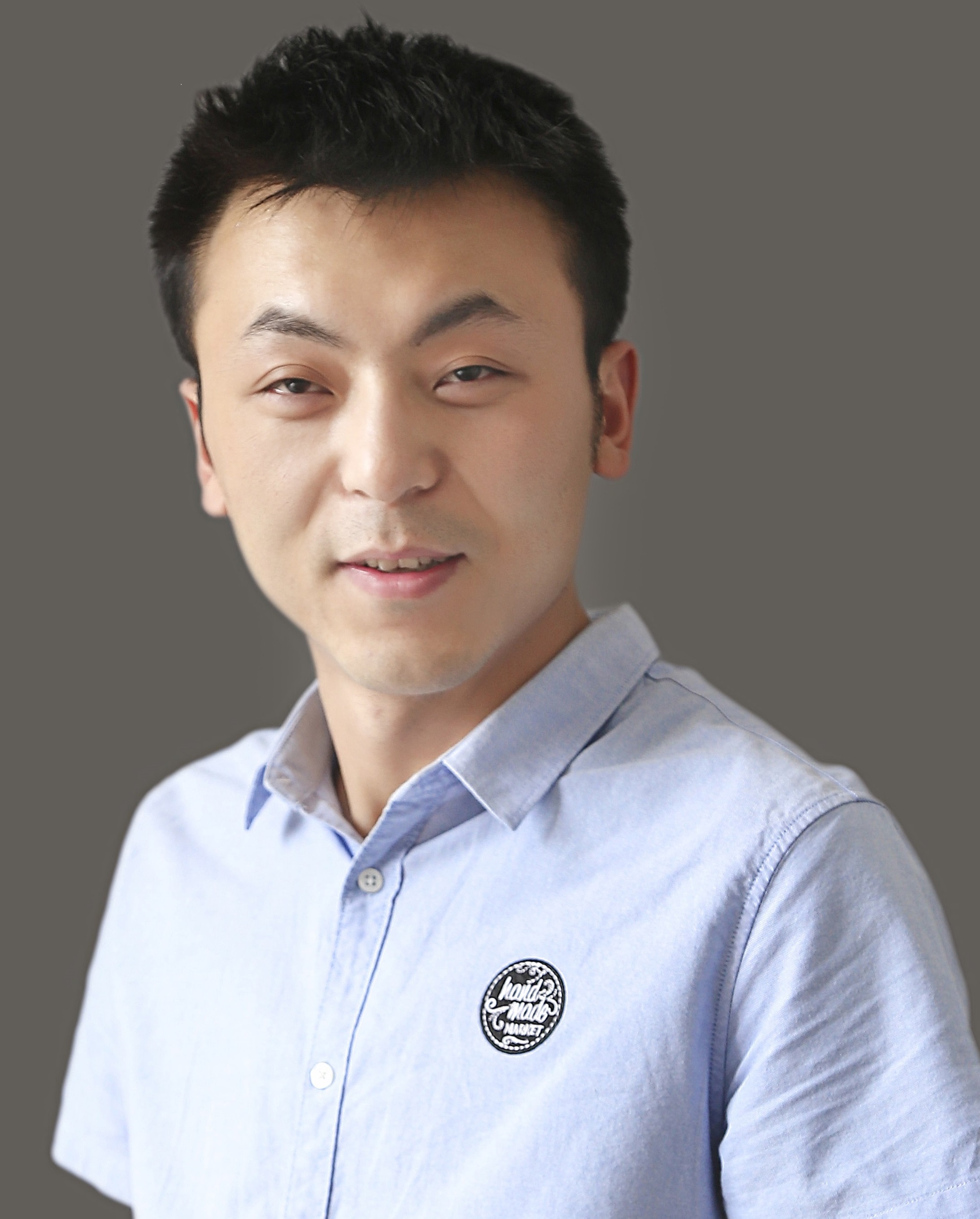}}]{Xu Liu}
(S'15) received the B.Sc. degrees in Mathematics and applied mathematics from North University of China, Taiyuan, China in 2013. He is currently pursuing the PhD degree in circuit and system from Xidian University, Xi'an China.
He is currently a member of Key Laboratory of Intelligent Perception and Image Understanding of Ministry of Education, International Research Center for Intelligent Perception and Computation, and Joint International Research Laboratory of Intelligent Perception and Computation. He is the chair of IEEE Xidian university student branch. Xidian University, Xi'an, China. His current research interests include machine learning, deep learning and image processing.
\end{IEEEbiography}

 \begin{IEEEbiography}[{\includegraphics[width=1in,height=1.25in,clip,keepaspectratio]{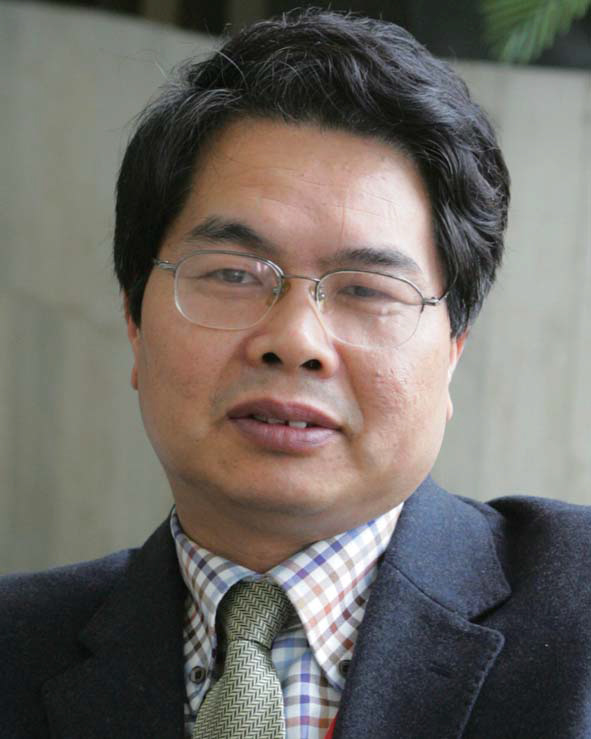}}]{Licheng Jiao}
 (F'17) received the B.S.degree from Shanghai Jiaotong University, Shanghai, China, in 1982 and the M.S. and PhD degree from Xi'an Jiaotong University, Xi'an, China, in 1984 and 1990, respectively.
Since 1992, he has been a Professor with the school of Electronic Engineering, Xidian University, Xi'an, where he is currently the Director of Key Laboratory of Intelligent Perception and Image Understanding of the Ministry of Education of China. His research interests include image processing, natural computation, machine learning, and intelligent information processing.
 \end{IEEEbiography}
 \begin{IEEEbiography}
 [{\includegraphics[width=1in,height=1.25in,clip,keepaspectratio]{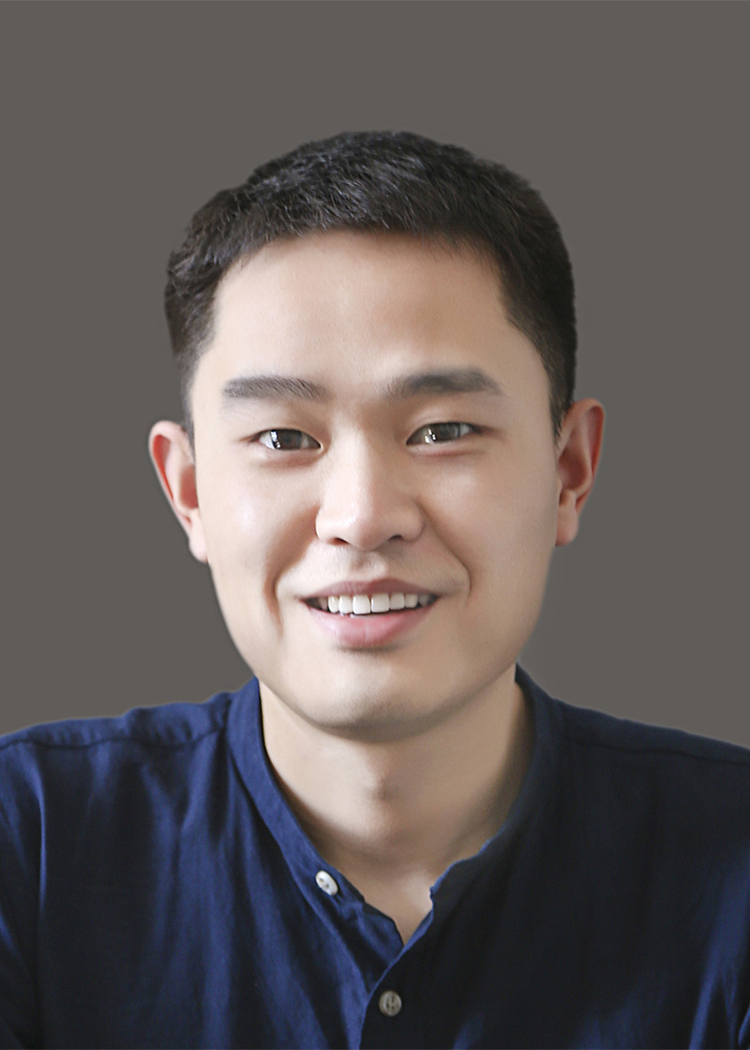}}]{Xu Tang}
 (M'17) received the B.S., M.S., and Ph.D. degrees from Xidian University, Xi'an, China, in 2007, 2010, and 2017, respectively, where he is currently pursuing the Ph.D. degree in circuit and systems.
He is currently a member with the Key Laboratory of Intelligent Perception and Image Understanding, Ministry of Education, Xidian University. His current research interests include remote sensing image processing, remote sensing image contentbased retrieval, and reranking.
 \end{IEEEbiography}

 \begin{IEEEbiography}
 [{\includegraphics[width=1in,height=1.25in,clip,keepaspectratio]{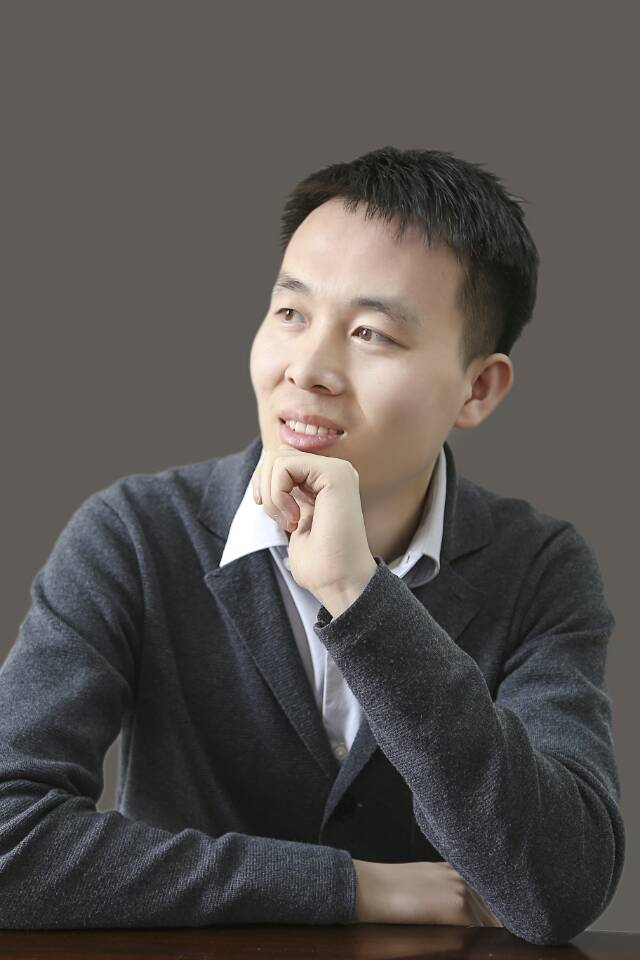}}]{Qigong Sun}
 (S'15) received the B.Eng. degrees in intelligence science and technology from Xidian University, Xi'an, China in 2015. He is currently pursuing the Ph.D. degree in circuit and system from Xidian University, Xi'an China. Currently, he is a member of Key Laboratory of Intelligent Perception and Image Understanding of Ministry of Education, and International Research Center for Intelligent Perception and Computation, Xidian University, Xi'an, China. His research interests include deep learning and image processing.
  \end{IEEEbiography}

 \begin{IEEEbiography}
 [{\includegraphics[width=1in,height=1.25in,clip,keepaspectratio]{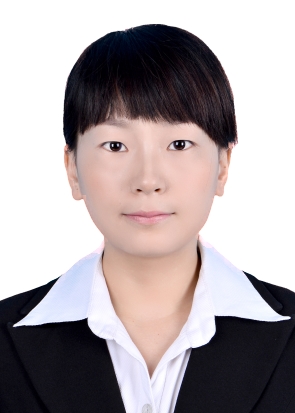}}]{Dan Zhang}
 received the B.S. degree in intelligent science and technology from Xidian University, Xi'an, China, in 2011. She was a Counselor with Xidian University, Xi'an, China, from 2011 to 2013. And she received the M.S. degree in circuits and systems from Xidian University, Xi'an, China, in 2016. She is currently working in the Key Laboratory of Intelligent Perception and Image Understanding of Ministry of Education, Xidian University.

Her current research interests include image processing, machine learning.

 \end{IEEEbiography}

%
%
%



\end{document}